\documentclass[10pt,twocolumn,letterpaper]{article}

\usepackage{cvpr}
\usepackage{times}
\usepackage{epsfig}
\usepackage{graphicx}
\usepackage{amsmath}
\usepackage{amssymb}

\usepackage[export]{adjustbox}
\usepackage{multirow}
\usepackage{array}
\usepackage{dsfont}
\usepackage{stfloats}
\usepackage{boldline}
\usepackage{hyphenat}

\def\VizImageWidth{0.668in}
\def\VizImageHeightLong{0.960in}
\def\VizImageHeightMiddle{0.600in}

\def\CurveHeight{1.25in}

\def\AttentionHeight{0.800in}
\usepackage[pagebackref=true,breaklinks=true,letterpaper=true,colorlinks,bookmarks=false]{hyperref}

\newcommand{\bst}[1]{\textcolor{red}{ \textbf{#1} }}
\newcommand{\second}[1]{\textcolor{blue}{ \textbf{#1} }}

\cvprfinalcopy 

\def\cvprPaperID{340} 

\ifcvprfinal\fi
\begin{document}

\title{FastMask: Segment Multi-scale Object Candidates in One Shot}

\author{Hexiang Hu\thanks{Equal contribution.} \ \thanks{Work was done during their internships at Megvii Inc.}\\
USC\\
{\tt\footnotesize hexiangh@usc.edu}
\and
Shiyi Lan\footnotemark[1] \ \footnotemark[2]\\
Fudan University\\
{\tt\footnotesize sylan14@fudan.edu.cn}
\and
Yuning Jiang\\
Megvii Inc.\\
{\tt\footnotesize jyn@megvii.com}
\and
Zhimin Cao\\
Megvii Inc.\\
{\tt\footnotesize czm@megvii.com}
\and
Fei Sha\\
USC\\
{\tt\footnotesize feisha@usc.edu}
}
\maketitle

\begin{abstract}
Objects appear to scale differently in natural images. This fact requires methods dealing with object-centric tasks (\eg object proposal) to have robust performance over variances in object scales. In the paper, we present a novel segment proposal framework, namely FastMask, which takes advantage of hierarchical features in deep convolutional neural networks to segment multi-scale objects in one shot. Innovatively, we adapt segment proposal network into three different functional components (body, neck and head). We further propose a weight-shared residual neck module as well as a scale-tolerant attentional head module for  efficient one-shot inference. On MS COCO benchmark, the proposed FastMask outperforms all state-of-the-art segment proposal methods in average recall being 2$\sim$5 times faster. Moreover, with a slight trade-off in accuracy, FastMask can segment objects in near real time ($\sim$13 fps) with 800$\times$600 resolution images, demonstrating its potential in practical applications. Our implementation is available on \url{https://github.com/voidrank/FastMask}. 
\end{abstract}



\section{Introduction}
\label{sec:intro}

\begin{figure}[htb]
\centering
\includegraphics[width=0.475\textwidth]{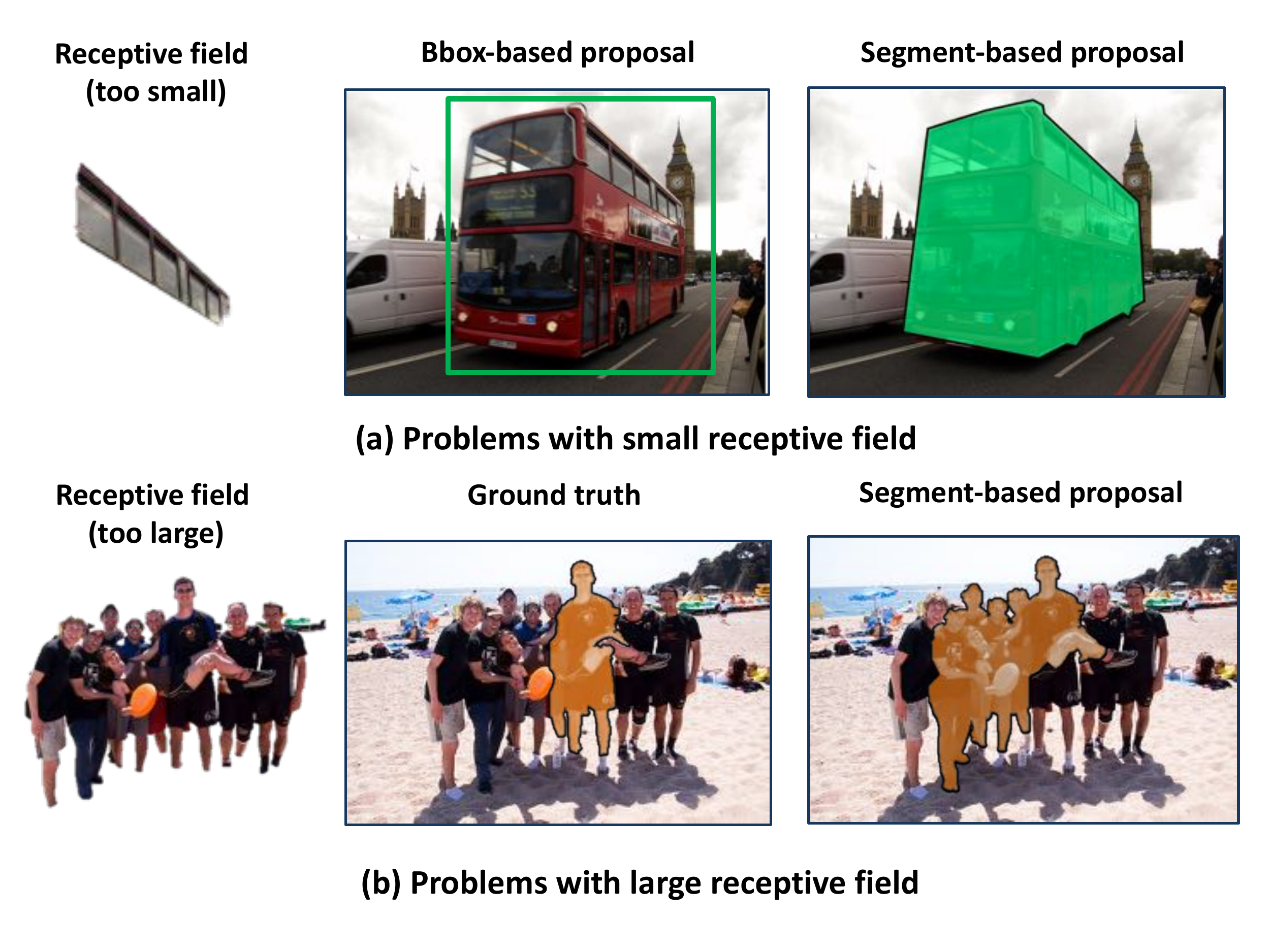}
\caption{How a mismatched receptive field affects the segment proposal results. Refer to text for detailed explanation. }
\vspace{-0.2in}
\label{fig:intro}
\end{figure}

Object proposal is considered as the first and fundamental step in object detection task~\cite{girshick_cvpr14,van_iccv11,arbelaez_cvpr14,krahenbuhl_eccv14,gokberk_cvpr13,zhang_cvpr11}. As the domain rapidly progressed, a renewed interest in object segment proposal has received intensive attentions~\cite{dai_cvpr16,pinheiro_nips15,pinheiro_eccv16,dai_eccv16,bell_cvpr16}. Different from traditional object proposal methods, segment proposal algorithms are expected to generate a pixel-wise segment instead of a bounding box for each object. From this perspective, segment proposal inherits from both object proposal and image segmentation, and takes a step further towards simultaneous detection and segmentation~\cite{habriharan_eccv14}, which brings more challenges to overcome.

Among all these challenges, how to tackle the scale variances in object appearance remains the most critical one. Compared to bounding-box-based (bbox-based) object proposal, scale variance becomes a more serious problem for segment proposal. It is due to that in segment proposal, a highly matched receptive field is demanded to distinguish the foreground object from background. In Figure~\ref{fig:intro} two examples are given to explain how a mismatched receptive field affects the segment proposal results:
on one hand (Figure~\ref{fig:intro}~(a)), when the receptive field of object proposer is much smaller than the object itself (\eg perceiving only a window of a bus), the bbox-based proposer could still roughly estimate the bounding box with prior knowledge. However, the mission becomes almost impossible for a segment-based proposer as they need to imagine the complete contour of the bus; on the other hand (Figure~\ref{fig:intro}~(b)), too large receptive field may introduce noises from backgrounds and result in the incorrect instance-level segments. For example, a segment-based proposer could be distracted by other people standing nearby the target person, leading to an inaccurate mask covering not only the target person. As a consequence, once the receptive field of a segment-based proposer is fixed, object scale variance will badly affect both segmentation fineness and proposal recall.

In general, existing methods~\cite{dai_cvpr16,pinheiro_nips15,pinheiro_eccv16,dai_eccv16,bell_cvpr16} could be divided into two major categories by how they deal with scale variances. The first category~\cite{dai_cvpr16,bell_cvpr16} uses extra bbox-based object proposals or object detections as initial inputs. However, its effectiveness and efficiency are highly dependent on the accuracy and speed of pre-processing proposal methods. The second one~\cite{pinheiro_nips15,pinheiro_eccv16,dai_eccv16} adopts the image pyramid strategy, in which the original image is rescaled and fed into a fixed-scale object proposer repeatedly for multi-scale inference (see Figure~\ref{fig:featurepyramid}(a)). However, such multi-shot methods face a common dilemma: a densely sampled image pyramid becomes the computational bottleneck of the whole framework; nevertheless, reducing the number of the scales of image pyramid leads to performance degradation. Such methods could hardly provide satisfactory accuracy and speed at the same time. With the observation that the original image has already contained all information of an image pyramid, we argue that using one single image should be enough to capture all multi-scale objects in it.

Therefore, in this paper, we aim to address the scale variances in segment proposal by leveraging the hierarchical feature pyramid\cite{girshick_cvpr15} from convolutional neural networks (CNN). We adapt segment proposal network into three different functional components, namely {\it body}, {\it neck} and {\it head}. Similar to~\cite{pinheiro_nips15,pinheiro_eccv16}, the body and head module are responsible for extracting semantic feature maps from original images and decoding segmentation masks from feature maps, respectively. Furthermore, We introduce the concept of neck module, whose job is to recurrently zoom out the feature maps extracted by the body module into feature pyramids, and then feed the feature pyramids into the head module for multi-scale inference. We summarize our main contributions as follows:

\begin{itemize}
\setlength{\itemsep}{0pt}
\setlength{\parskip}{0pt}
\setlength{\parsep}{0pt}
\vspace{-0.05in}
  \item First, we learn a novel weight-shared residual neck module to build a feature pyramid of CNN while preserving a well-calibrated feature semantics, for efficient multi-scale training and inference.
      \vspace{0.02in}
  \item Next, we propose a novel scale-tolerant head module which takes advantage of visual attention and significantly reduces the impact of background noises caused by mismatched scales in receptive fields.
      \vspace{0.02in}
  \item Finally, together with all those modules, we make a framework capable of one-shot segment proposal. We evaluate our framework on MS COCO benchmark~\cite{lin_eccv14} and it achieves the state-of-the-art results while running in near real time.
\end{itemize}


\section{Related Work}
\label{sec:relatedwork}

\noindent{\textbf{Bbox-based object proposal.}} Most of the bbox-based object proposal methods rely on the dense sliding windows on image pyramid. In EdgeBox~\cite{zitnick_eccv14} and Bing~\cite{cheng_cvpr14}, the edge feature is used to make the prediction for each sliding window while the gradient feature is used in~\cite{zhang_cvpr11}. More recently, DeepBox~\cite{kuo_iccv15} trains a CNN to re-rank the proposals generated by EdgeBox, while MultiBox~\cite{erhan_cvpr14} generates the proposals from convolutional feature maps directly. Ren et. al.~\cite{ren_nips15} presented a region proposal network (RPN) is proposed to handle object candidates in varying scales.

\noindent{\textbf{Segment-based object proposal.}} Segments proposal algorithms aim to find diverse regions in an image which are likely to contain objects. Traditional segment proposal methods such as SelectiveSearch~\cite{van_iccv11}, MCG~\cite{arbelaez_cvpr14} and Geodesic~\cite{krahenbuhl_eccv14} first over-segment image into super pixels and then merge the super pixels in a bottom-up fashion. Inspired by the success of CNNs in image segmentation~\cite{shelhamer_pami16,chen_iclr15,yu_iclr16}, previous works~\cite{dai_cvpr16,bell_cvpr16} perform segmentation on the bbox-based object proposal results to obtain object segments. As the state-of-the-arts, DeepMask~\cite{pinheiro_nips15} proposes a body-head structure to decode object masks from CNN feature maps, and SharpMask~\cite{pinheiro_eccv16} further adds a backward branch to refine the masks. However, all these methods rely on an image pyramid during inference, which limits their application in practice.

\noindent{\textbf{Visual attention.}}
Instead of using holistic image feature from CNN, a number of recent works~\cite{xu_icml15,liu_arxiv2016,zhu_cvpr16,yang_cvpr2016} have explored visual attention to highlight discriminative region inside images and reduce the effects of noisy background. In this paper we apply such attention mechanism to improve the instance-level segmentation performance.

\begin{figure*}[bp]
\centering
\includegraphics[width=1\textwidth]{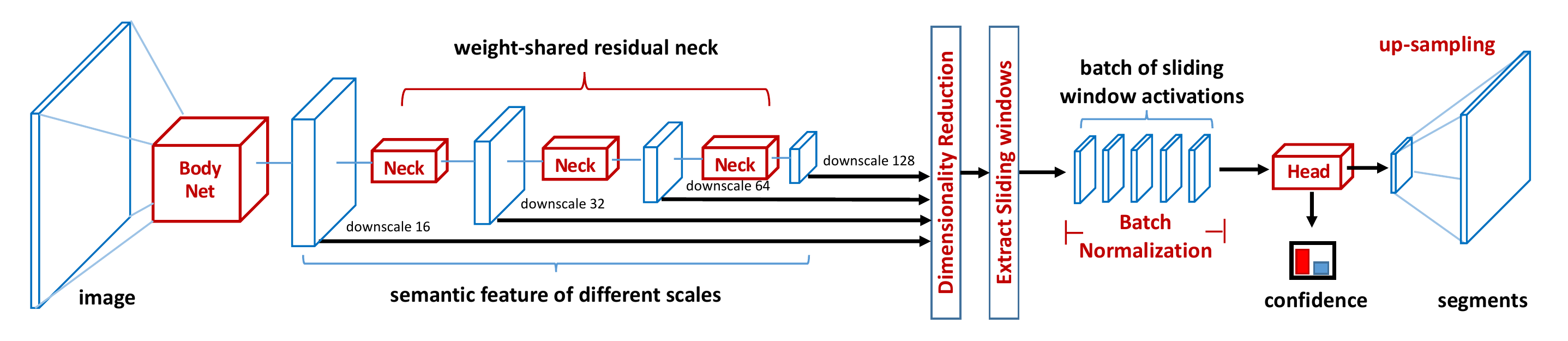}
\caption{An overview of the proposed FastMask architecture.}
\vspace{-0.1in}
\label{fig:architecture}
\end{figure*}

\section{From Multi-shot to One-Shot}
\label{sec:reviewofdeepmask}

DeepMask~\cite{pinheiro_nips15} is considered as the representative of the CNN-based multi-shot segment proposal methods, where a body-head structure is proposed. In this section, we briefly review DeepMask to help better understand the multi-shot paradigm and then proceed to our proposed one-shot paradigm.

\noindent{\textbf{Patch-based training.}} DeepMask is trained to predict a segmentation mask and a confidence score given a fixed-size image patch. In training, an image patch is assigned to be positive if it satisfies the \textit{object-centric constrain}~\cite{pinheiro_nips15}; otherwise negative. All the image patches are cropped and rescaled into fixed size (\eg 224$\times$224). These patches are fed into the body network of DeepMask to extract semantic feature maps, and then decoded into the confidence scores and the segmentation masks using the head module.

\begin{figure}[tp]
\centering
\includegraphics[width=0.485\textwidth]{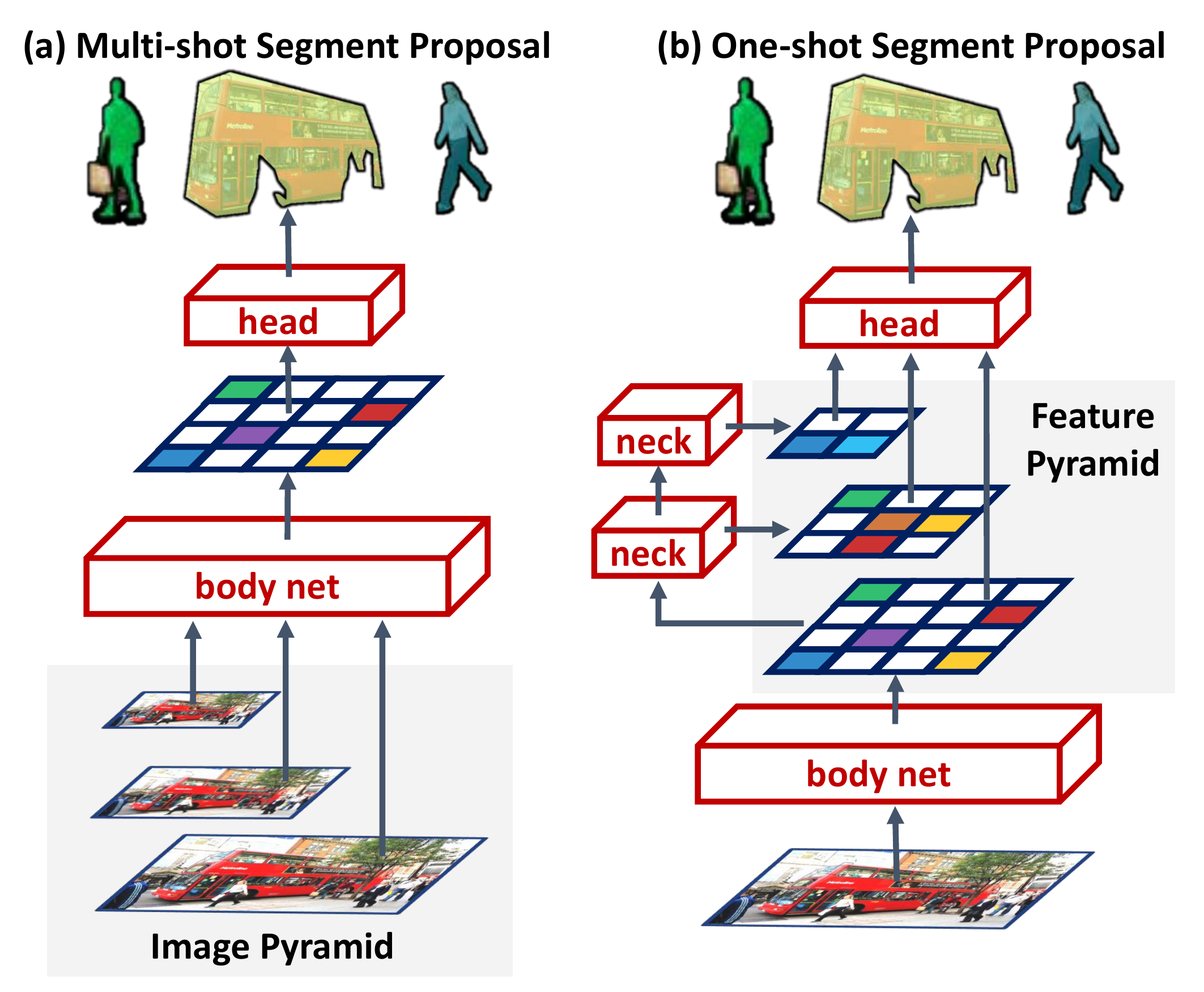}
\caption{ Comparison between multi-shot paradigm and our one-shot paradigm.}
\vspace{-0.2in}
\label{fig:featurepyramid}
\end{figure}

\noindent{\textbf{Multi-shot inference.}} During multi-shot inference, DeepMask applies the trained model densely at each location, repeatedly across different scales. As shown in Figure~\ref{fig:featurepyramid}~(a), at first the input image is resized repeatedly into an image pyramid. Next, the body network of DeepMask extracts a full feature map from each resized image. Finally the head module is applied on every fixed-size sliding window (e.g., 14$\times$14) on multi-scale feature maps, to decodes the confidence score and mask for each sliding window.

For DeepMask and its variants\cite{pinheiro_nips15,pinheiro_eccv16,dai_cvpr16}, a densely sampled image pyramid is required during inference. However, as the convolutional computation over image pyramid is redundant, the image pyramid has become the computational bottleneck in such multi-shot segment proposal methods.

To overcome the inefficiency brought by image pyramid, we propose a one-shot paradigm that enables efficient training and inference. As shown in Figure~\ref{fig:featurepyramid}~(b), we inherit the body-head structure and introduce a new component called {\it neck}. This neck component could be used on the feature map and zoom it out into feature pyramid while preserving feature semantics. Then, a shared head module is applied on the pyramid of feature maps to decode object segments at different scales. With the proposed body-neck-head structure, we could save the redundant convolutional computation and make efficient use of information to perform segment proposal in one shot. We refer this as \textit{\textbf{one-shot segment proposal paradigm}} and derive our proposed segment proposal framework in Section~\ref{sec:method}.

\section{Our Approach}
\label{sec:method}

In this section, we introduce our approach in detail. First, we overview the proposed architecture (FastMask), to give a concrete idea about our body-neck-head structure. We explain our entire pipeline by illustrating the data flow from input image to object segments. Next we study the different designs of the neck module, including both the non-parametric and parametric necks. Finally, we present a novel head module that enables scale-tolerant segmentation mask decoding by taking advantage of the attention model, which plays the key role in improving performance.

\subsection{Network Architecture }
\label{sec:method:architecture}

We present our network architecture in Figure~\ref{fig:architecture}. Similar to multi-shot methods, the body network extracts semantic feature from the input image. With this base feature map, a shared neck module is applied recursively at it to build feature maps with different scales. This pyramid of feature maps are then input to a $1 \times 1$ convolution for reducing dimensionality. Next, we extract dense sliding windows from all these feature maps, and do a batch normalization across all windows to calibrate window features. Note that with a feature map downscaled by a factor \textit{m}, a sliding window of size (\textit{k}, \textit{k}) corresponds to a patch of (\textit{m} $\times$  \textit{k}, \textit{m} $\times$ \textit{k}) at original image. Finally, a unified head module is used to decode these sliding-window features and produce the output confidence score as well as object mask.

Our approach could be easily adopted to any existing CNN architectures (\eg VGGNet~\cite{simonyan_iclr15}, ResNet~\cite{he_cvpr16}), by replacing their fully connected layers or some convolutional and pooling layers on the top with the neck and head modules. The reason for removing those top convolutional and pooling layers is to keep feature map in a feasible size, so that a small object could still correspond to a notable region on feature map.

\subsection{Residual Neck}
\label{sec:method:neck}

\begin{figure}[tp]
\centering
\includegraphics[width=0.49\textwidth]{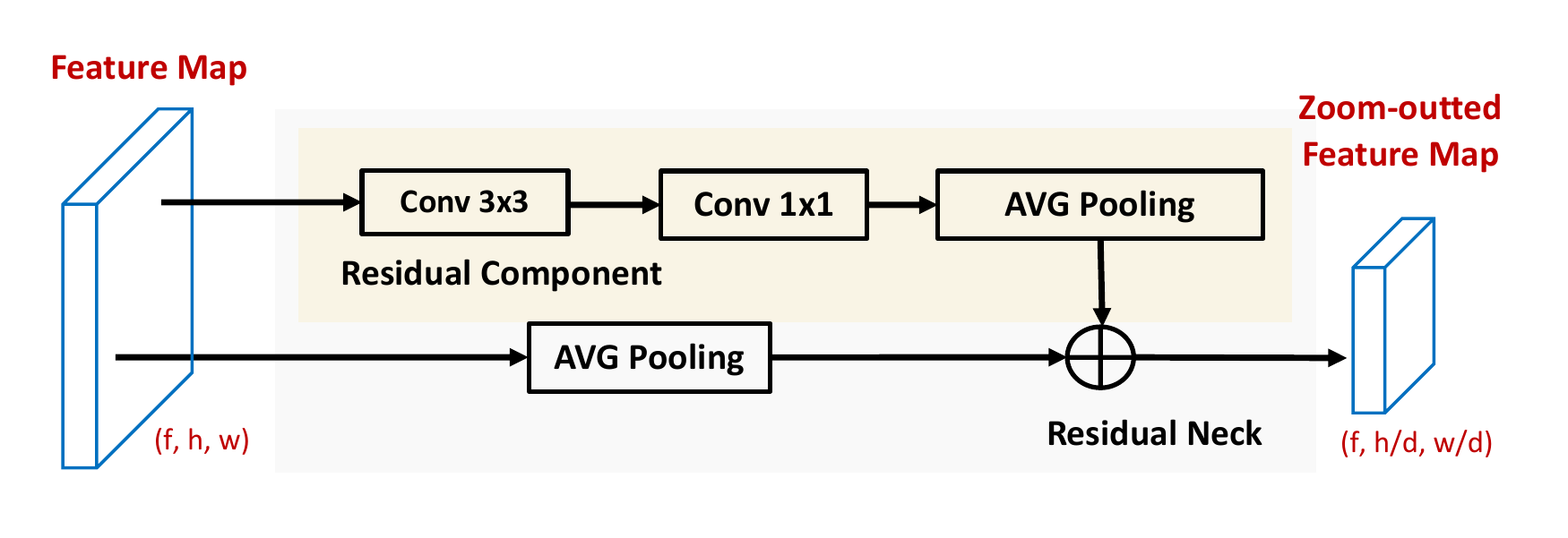}
\vspace{-0.2in}
\caption{ Illustration of the residual neck. We augment the average pooling neck with a learnable residual component.}
\vspace{-0.1in}
\label{fig:method:resneck}
\end{figure}

\begin{figure*}[bp]
\centering
\vspace{-0.1in}
\includegraphics[width=1\textwidth]{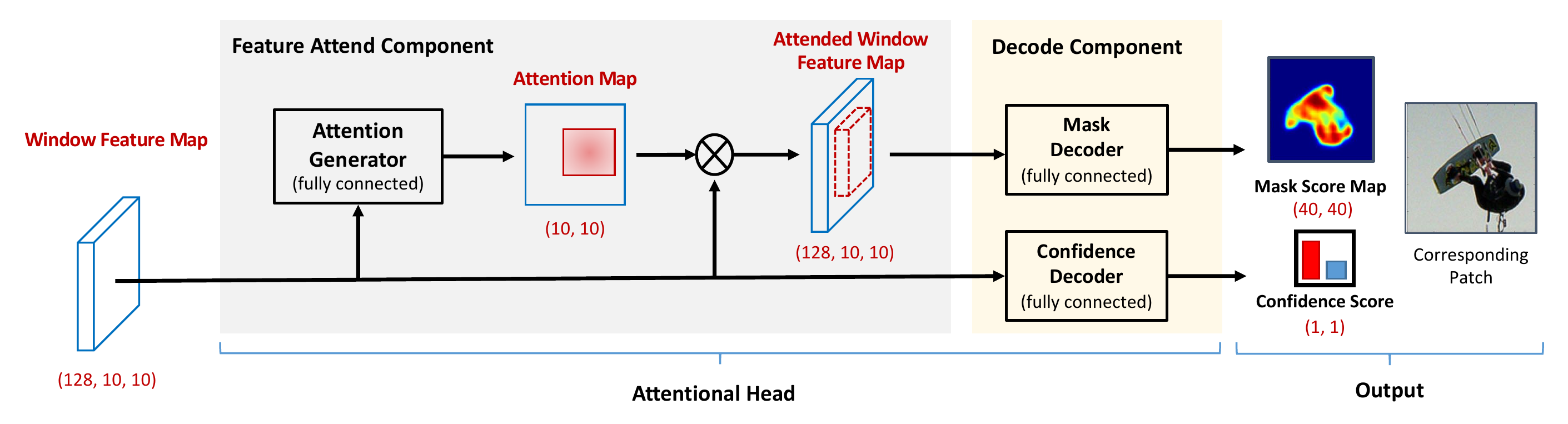}
\caption{ Details of the attentional head. It presents the data flow starting from a feature map to the confidence score and segment mask inside each sliding window. (Notations in round brackets indicate the dimensionality) }
\vspace{-0.1in}
\label{fig:method:head}
\end{figure*}

We consider both non-parametric and parametric methods for encoding feature pyramid. To zoom out feature map, a straightforward choice is non-parametric pooling. Both max pooling and average pooling are widely used components in modern CNN architectures on recognition and detection. In our scenario, we would like to calibrate each feature map for a unified decoding. However, some pooling necks generate sub-optimal empirical results as desired by their natural. In this section, we discuss about several choices of the necks and compare them empirically.

\noindent{\textbf{Max pooling neck.}}
Max pooling produces uncalibrated features during encoding. With spatial grids of feature, max pooling takes the max response over each grid for downscaled feature maps. As a result, this process increases the mean of output feature maps. As max pooling is repeatedly applied, the top feature maps would have significantly larger mean than bottom ones.

\noindent{\textbf{Average pooling neck.}}
Average pooling smooths out discriminative feature during encoding. Different from max pooling, average pooling maintains the mean of feature maps. Although it helps to keep the means of features in different scales calibrated, it blurs discriminative feature. The lost of discriminative feature makes the head module suffer from distinguishing the object to its background.

\noindent{\textbf{Feed-forward neck.}}
To alleviate above side-effects, we propose to learn parametric necks that preserve feature semantics. One naive parametric choice is to learn a feed-forward neck which uses convolutional and pooling layers to zoom out feature maps. However, the feed-forward neck faces the gradient vanishing effect~\cite{he_eccv16} as the number of scales increases. In addition, feature semantics may change substantially since the feature maps on the top go through more convolutional operations than the bottom ones.

\noindent{\textbf{Residual neck.}}
Inspired by bottle-neck connection in~\cite{he_cvpr16}, we design to learn a residual neck as in Figure~\ref{fig:method:resneck}. We augment the non-parametric average pooling with a parametric residual component (using the same structure as in the feed-forward neck, a $3\times3$ convolutional layer followed by a $1\times1$ one) to zoom out feature maps, in order to reduce the the smooth effect of average pooling as well as preserve feature semantics.

\noindent{\textbf{Comparison.}}
To verify the effectiveness of the proposed necks, we empirically evaluate all these designs and report their performance in Table~\ref{tab:eval:neck}. Here we report overall AR@100 and AR@100 for objects in different sizes (details in Section~\ref{sec:experiment}). The results confirm that the residual neck component beats all other necks in terms of average recall. Note that we obtain a large margin in average recall for objects in large scale, which are decoded from the top feature maps. This verifies the effectiveness of the residual neck in encoding feature pyramid.

\subsection{Attentional Head}
\label{sec:method:head}

\begin{table}[tp]
\small
\tabcolsep 3pt
\centering
\begin{tabular}{ c|cccc }
Method            & AR@100 & AR$^S$@100 & AR$^M$@100 & AR$^L$@100 \\ \hlineB{3}
{ Avg-Pooling }   & 27.9 & 11.5 & 36.9 & 43.9 \\
{ Max-Pooling }   & 27.8 & 11.1 & 36.8 & 44.2 \\
{ Feed-Forward }  & 27.1 & 10.8 & 35.8 & 43.4 \\
{ Residual }      & \bst{29.3} & \bst{11.7} & \bst{38.3} & \bst{47.2} \\
\hline
\end{tabular}
\vspace{0.1in}
\caption{Comparison on different designs of the neck modules (on COCO benchmark). VGGNet~\cite{simonyan_iclr15} is used as body network for all the necks.}
\label{tab:eval:neck}
\end{table}

Following~\cite{pinheiro_nips15, pinheiro_eccv16}, we use a combination of convolutional layers and fully connected layers to assemble a head module for decoding mask and object confidence. However, in the context of feature pyramid decoding, we found that simply applying this head leads to a suboptimal performance. A likely reason is that, comparing to original DeepMask~\cite{pinheiro_nips15}, our feature pyramid is sparser in scales. To be concrete, after the neck module is applied, the feature map is downscaled by a factor of two, which means that the scale gap between two adjacent feature maps is two (while the scale gap in DeepMask is $2^{0.5}$). The sparse feature pyramid raises the possibility that no suitable feature maps exists for an object to decode, and also increases the risk of introducing background noises because an object may not matches well with the size of receptive field (sliding window).

Such observations drive us to propose two alternative solutions alleviating such problem. First, we tried to expand our network into two stream, to simply increase the scale density (we defer this part to Section~\ref{sec:implementation}). Second, we develop a novel head module that learns to attend salient region during decoding. With visual attention, a decoding head could reduce the noises from the backgrounds in a sliding window and alleviate the mismatch between the size of receptive field and object. Note that such attention also brings the tolerance to shift disturbance (\ie when a object is not well centered), which further improves its robustness.

Figure~\ref{fig:method:head} gives the detailed implementation of our attentional head. Given the feature map of a sliding window as input, we first compute a spatial attention through a fully connected layer. This spatial attention is then applied to window feature map via an element-wise multiplication across channels. Such operation enables the head module to highlight features on the salient region, which indicates the rough location for the target object. Finally, this attended feature map is input into a fully connected layer to decode the segmentation mask of the object.

\begin{table}[bp]
\small
\tabcolsep 8pt
\centering
\begin{tabular}{ c|ccc }
Method                  & AR@10      & AR@100     & AR@1k \\ \hlineB{3}
{Standard Head}         & 12.7       & 24.8       & 33.2 \\
{Attentional Head}      & \bst{15.2} & \bst{29.3} & \bst{38.6} \\
\hline
\end{tabular}
\vspace{0.1in}
\caption{Comparison of different head modules on the COCO benchmark. VGGNet~\cite{simonyan_iclr15} is used as the body network.}
\label{tab:eval:head}
\end{table}

\begin{figure}[bp]
\footnotesize
\centering
\tabcolsep 8pt
\begin{tabular}{ccc}

\textbf{\footnotesize Sliding Window} & \textbf{\footnotesize Attention Pred.} & \textbf{\footnotesize Segment Pred.} \\

\includegraphics[height=\AttentionHeight,frame]{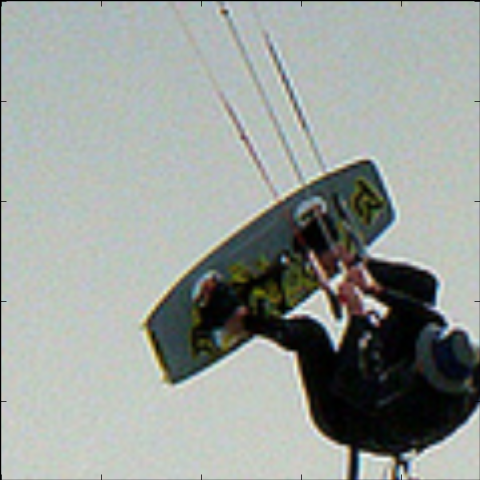} &
\includegraphics[height=\AttentionHeight,frame]{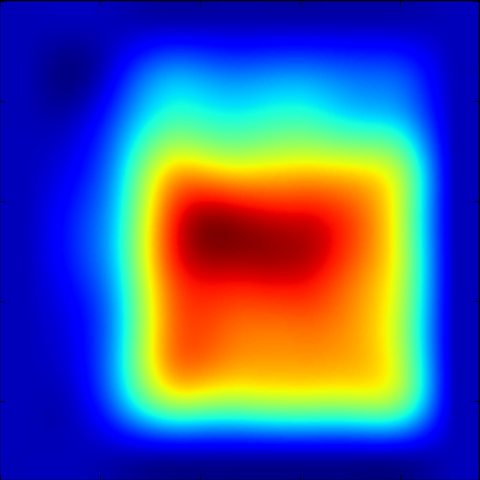} &
\includegraphics[height=\AttentionHeight,frame]{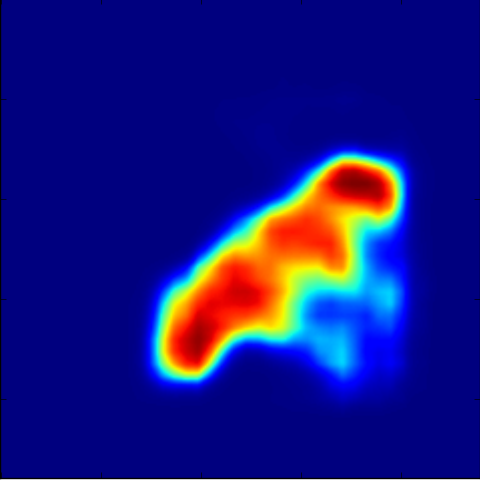} \\

\includegraphics[height=\AttentionHeight,frame]{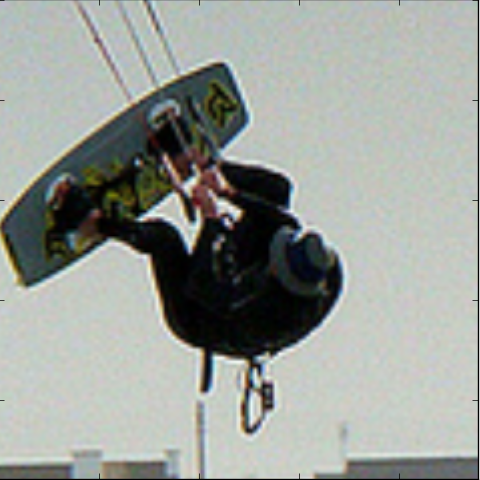} &
\includegraphics[height=\AttentionHeight,frame]{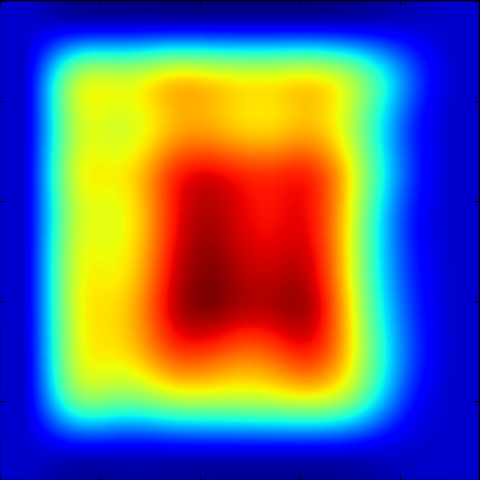} &
\includegraphics[height=\AttentionHeight,frame]{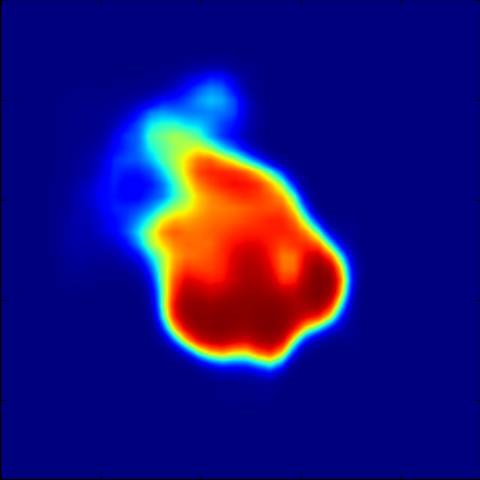} \\

\end{tabular}
\caption{ \textbf{Attention exemplars.} Attentional head helps to locate important central object feature for mask decoder. (Color towards red represents high score)}
\vspace*{-0.1in}
\label{fig:eval:attention}
\end{figure}

\noindent{\textbf{Comparison.}}
To verify the effectiveness of the proposed attentional head, we do experimental comparisons between FastMask with a standard head and FastMask with an attentional head, as reported in Table~\ref{tab:eval:head}. From the table we can see that with the tolerance to scale and shift disturbance, the attentional head significantly improves the segment proposal accuracy.

\noindent{\textbf{Visualization.}}
To further justify the effectiveness of regional attention in denoising, we visualize two examples (Figure~\ref{fig:eval:attention}) as exemplars. In the top example, a skateboard is the central object and the person riding it is the noisy. As a consequence, generated attention weight regions close to skateboard with higher confidence to highlight the central object. Similarly, the bottom example indicates the same spirit, while in a vice versus manner that person becomes the central object and skateboard is the noise.

\section{ Implementation Details }
\label{sec:implementation}
 In this section we first present an practical technique for obtaining more scales in feature pyramid. Then we give all the details about training, optimization and inference in our framework. We made our code public available on: \url{https://github.com/voidrank/FastMask}.

\begin{figure*}[tp]
\centering
\includegraphics[width=1\textwidth]{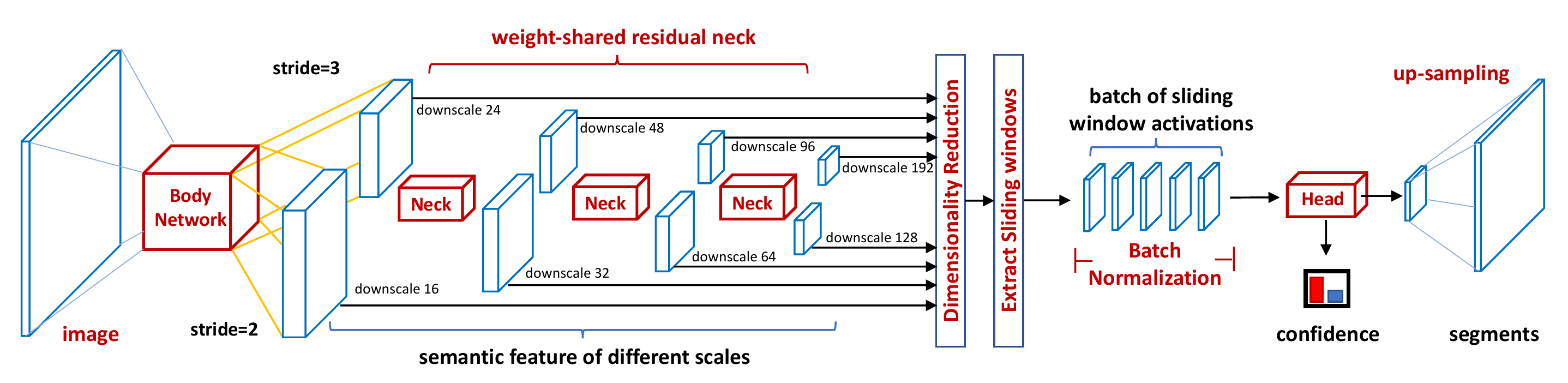}
\caption{An overview of the proposed two-stream FastMask architecture.}
\vspace{-0.1in}
\label{fig:impl:architecture-two-stream}
\end{figure*}

\subsection{Two-stream network}
\label{sec:implementation:two-stream}
As mentioned in Section~\ref{sec:method:head}, to make the feature pyramid denser, we craft the body network (Shown in Figure~\ref{fig:impl:architecture-two-stream}) to branches in the middle through applying pooling layers with different strides (\eg 2 and 3 in our implementation) and feed these differently scaled features to the shared neck. It augments the body network with capability to produce features of diverse sizes, not necessarily limited to a multiple of two.

In our practice, we branch a 2 $\times$ 2 pooling on the feature downscaled by 8 to generate feature downscaled by factors of 16 and 24, and input these feature to the shared top convolutions. Then we apply our neck and head modules on these two streams to produce object segments in different scales. This technique adds more scales of feature, helps FastMask to be more robust to scale difference, but introduce limited extra computation. Note that we do not add any new parameters for learning through this branching technique.

\subsection{ Training }
\label{sec:implementation:train}
The key difference between training FastMask and standard DeepMask~\cite{pinheiro_nips15} is that FastMask could be trained by images in varying scales, rather than cropped fixed-scale patches. To enable this training scheme, we introduce our strategies on ground truth assignment, learning objective and optimization details.

\noindent{\textbf{Ground truth assignment}}.
During training, we need to determine which sliding window a ground truth object belongs to. For each ground truth object, we assign it to a sliding window if \textit{(i)} it fully contains this object, and \textit{(ii)} the object fits into the scale range of [0.4, 0.8] with regard to the window, and \textit{(iii)} the object is roughly centered in the window (object center in central 10$\times$10 rectangle region of window). Once an object is assigned to a window, we extract the segmentation mask as segmentation ground truth (denoted by $s$) and use the surrounding bounding as attention ground truth (denoted by $a$).

\noindent{\textbf{Learning objective}}.
The overall objective function of FastMask is a weighted sum of the confidence loss ($\mathcal{L}_{conf}$), segmentation loss (\textit{$\mathcal{L}_{seg}$}) and region attention loss ($\mathcal{L}_{att}$). Note that $c$, $a$, $s$ stand for ground truth label for confidence, region attention and segmentation mask, while $\hat{c}$, $\hat{a}$, $\hat{s}$ stand for corresponding prediction.

\begin{eqnarray}
\vspace{-0.2in}
\label{eqn:loss:overall}
&\mathcal{L}(c, a, s) = \frac{1}{N} \sum^N_k \big[ \mathcal{L}_{conf}(c_k, \hat{c}_k) \nonumber \\
&+ \mathds{1}(c_k) \cdot \big( \mathcal{L}_{seg}(s_k, \hat{s}_k) + \mathcal{L}_{att}(a_k, \hat{a}_k) \big) \big].
\end{eqnarray}

Here $\mathds{1}(c_k)$ is an indicator function which returns 1 if $c_k$ is true and 0 otherwise. Equation \ref{eqn:loss:overall} indicates that we only back-propagate gradients when $c_k = 1$. It is critical to get good performance by computing $\mathcal{L}_{seg}$ and $\mathcal{L}_{att}$ only with positive object samples. We normalize this weighted sum with the total number of sliding windows across mini-batches. For each loss components, we compute the cross entropy function between the prediction and ground truth as following:

\begin{eqnarray}
\vspace{-0.2in}
&\mathcal{L}_{conf}(c, \hat{c}) &= - \mathcal{E}(s_{i,j}, \hat{s}_{i,j}) \label{eqn:loss:conf} \\
&\mathcal{L}_{seg}(s, \hat{s}) &= - \big[ \frac{1}{w \cdot h} \sum^{h, w}_{i,j} \big( \mathcal{E}(s_{i,j}, \hat{s}_{i,j})  \big)  \big] \label{eqn:loss:seg} \\
&\mathcal{L}_{att}(a, \hat{a})  &= - \big[ \frac{1}{w \cdot h} \sum^{h, w}_{i,j} \big(  \mathcal{E}(a_{i,j}, \hat{a}_{i,j}) \big)  \big] \label{eqn:loss:att}.
\end{eqnarray}

For $\mathcal{L}_{seg}$ and $\mathcal{L}_{att}$, we normalize spatially across the window to balance the gradients between three loss components. $\mathcal{E}(y, \hat{y})$ is a standard binary cross entropy function with sigmoid activation function (denoted by $\sigma(y)$), in the following form:

\begin{eqnarray}
\vspace{-0.2in}
\mathcal{E}(y, \hat{y}) &= y \cdot log(\sigma{(\hat{y})}) + ( 1 - y) \cdot log(1 - \sigma{(\hat{y})} ) \label{eqn:loss:cross-entropy}.
\end{eqnarray}

\begin{table*}[tp]
\small
\tabcolsep 5pt
\centering
\begin{tabular}{ cc|ccc|ccc }
&& \multicolumn{3}{c}{\textbf{Box Proposals}} & \multicolumn{3}{c}{\textbf{Segmentation Proposals}} \\
Method & Body Net & AR@10 & AR@100 & AR@1k & AR@10 & AR@100 & AR@1k \\
\hlineB{3}
{MCG}                                       & - & 10.1 & 24.6 & 39.8 & 7.7 & 18.6 & 29.9        \\
{DeepMask~\cite{pinheiro_nips15}}           & VGG & 15.3 & 31.3 & 44.6 & 12.6 & 24.5 & 33.1     \\
{DeepMaskZoom~\cite{pinheiro_nips15}}       & VGG & 15.0 & 32.6 & 48.2 & 12.7 & 26.1 & 36.6     \\
{DeepMask$^*$~\cite{pinheiro_eccv16}}       & Res39 & 18.0 & 34.8 & 47.0 & 14.1 & 25.8 & 33.6   \\
{SharpMask~\cite{pinheiro_eccv16}}          & Res39 & 19.2 & 36.2 & 48.3 & 15.4 & 27.8 & 36.0   \\
{SharpMaskZoom~\cite{pinheiro_eccv16}}      & Res39 & 19.2 & 39.0 & 53.2 & 15.6 & 30.4 & \second{40.1} \\
{InstanceFCN~\cite{dai_eccv16}}             & VGG & - & - & - & 16.6 & \bst{31.7} & 39.2 \\
\hline
{FastMask+two streams}                      & Res39 & \second{22.6} & \second{43.1} & \bst{57.4} & \second{16.9} & \second{31.3} & \bst{40.6} \\
{FastMask+two streams}                      & Pva   & \bst{24.1} & \bst{43.6} & \second{56.2} & \bst{17.5} & 30.7 & 39.0 \\

\hline
\end{tabular}
\vspace{0.1in}
\caption{Object segment proposal results on COCO validation set for box and segmentation proposals. Note that we also report the body network for each corresponding method. }
\vspace{-0.1in}
\label{tab:exp:coco}
\end{table*}

\noindent{\textbf{Optimization}}.
We optimize the objective by standard stochastic gradient descent (SGD) with batch size equals 1, momentum equals 0.9 and weight decay equals 0.00005. We train our network for approximately 15 epochs and choose the best models through a different subset of COCO validation set. Following the practice of ~\cite{ren_nips15, girshick_cvpr14}, we balance positive and negative samples by a certain ratio (\eg roughly 1:1 in our case) after collecting all sliding-windows in training. In our practice, due to the limitation of GPU Memory, we train our two-stream network with totally 7-scale feature maps, by taking zooming out 4 times on the stream with $\textit{stride}=2$, and 3 times on the stream with $\textit{stride}=3$.

\subsection{ Inference }
\label{sec:implementation:inference}

During inference, we process an image in one shot and extract windows at multi-scale feature maps as same as the training stage. First the confidence score of each window is predicted, and then only the top-$\textit{k}$ confident windows are selected for object segment decoding. In addition, as the residual neck is weight shared, we could add or reduce the number of neck components during inference. This enables us to make easy trade-off between the effectiveness and efficiency, via adjusting the number of neck components. Therefore, although trained by 7 scales, the two-stream network could still be equipped by more than 7 neck modules to generate a denser feature pyramid. In the following experiments, unless specified, we use the two-stream network with 8 scales in the inference stage.

\section{ Experiments }
\label{sec:experiment}

\begin{figure*}[tp]
\footnotesize
\centering
\tabcolsep 20pt
\begin{tabular}{ccc}

\textbf{\footnotesize(a) Recall@10 Box Proposals} & \textbf{\footnotesize(b) Recall@100 Box Proposals} & \textbf{\footnotesize(c) Recall@1000 Box Proposals} \\

\includegraphics[trim=50 20 50 20,clip,height=\CurveHeight]{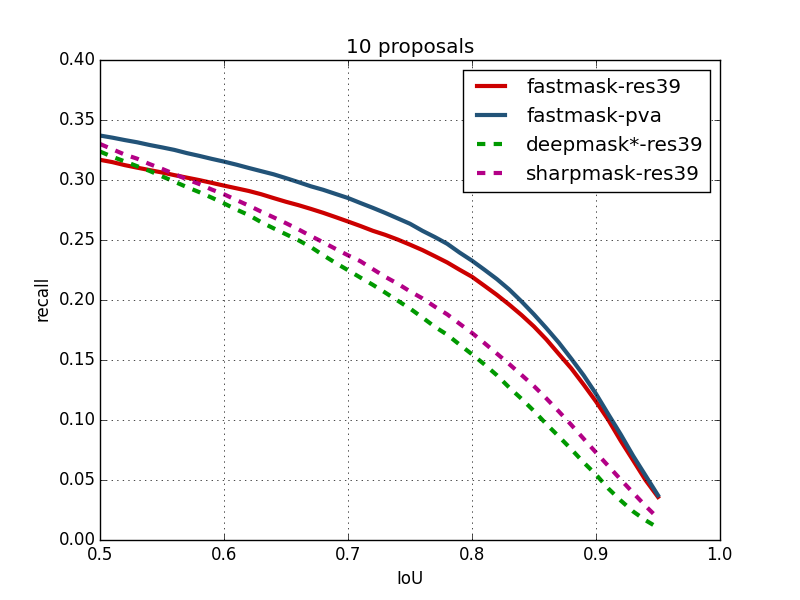} &
\includegraphics[trim=50 20 50 20,clip,height=\CurveHeight]{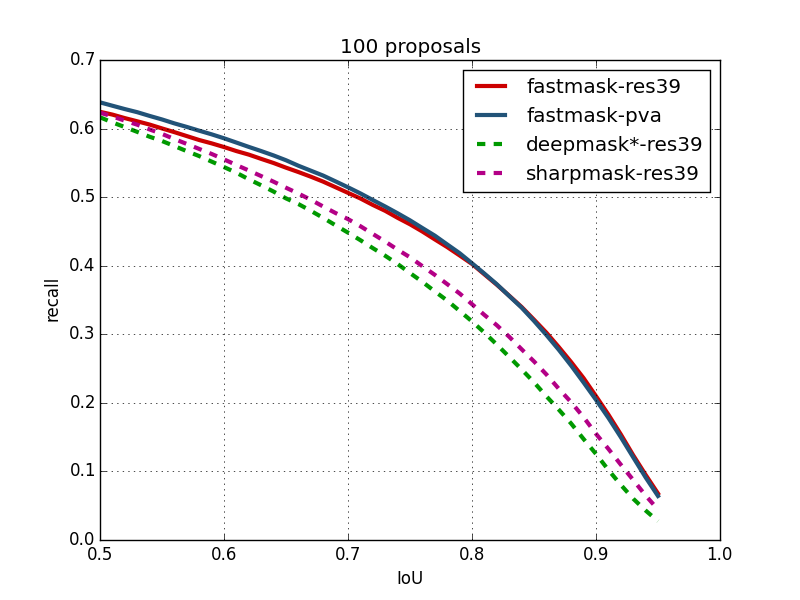} &
\includegraphics[trim=50 20 50 20,clip,height=\CurveHeight]{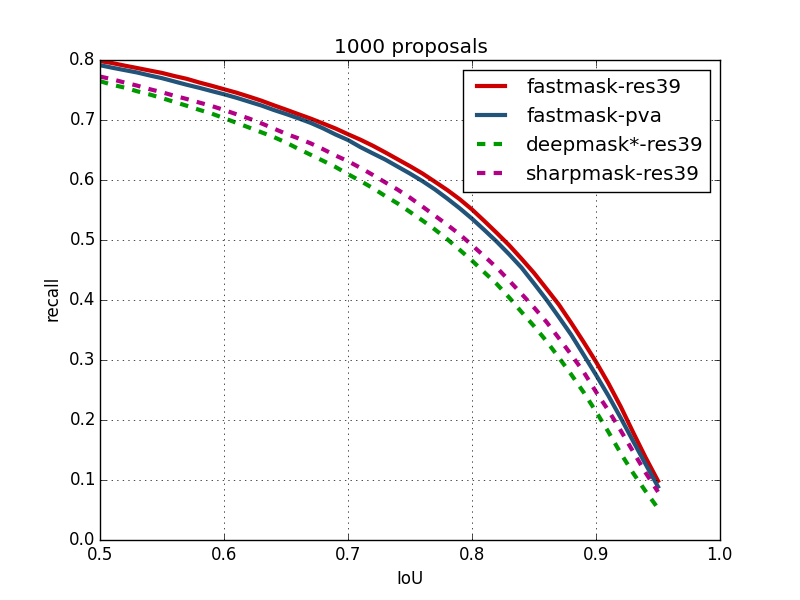} \\

\textbf{\footnotesize(d) Recall@10 Segment Proposals} & \textbf{\footnotesize(e) Recall@100 Segment Proposals} & \textbf{\footnotesize(f) Recall@1000 Segment Proposals} \\
\includegraphics[trim=50 20 50 20,clip,height=\CurveHeight]{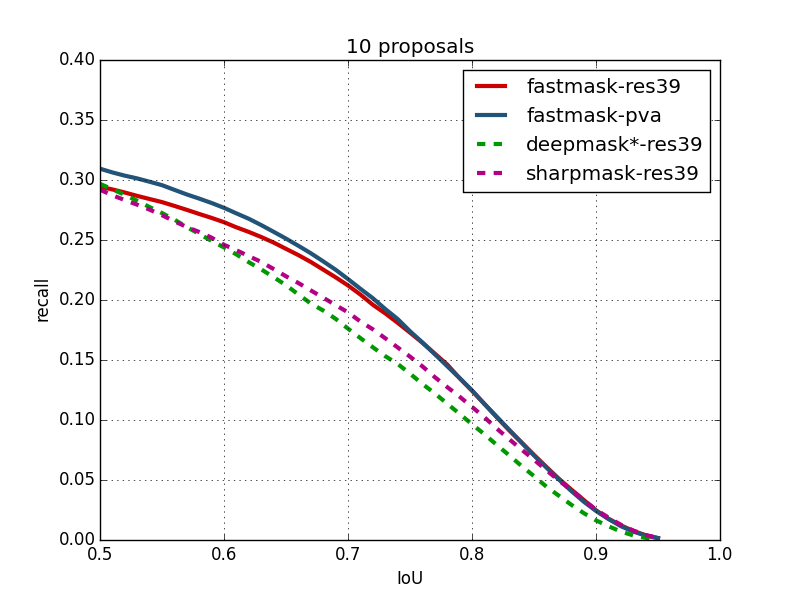} &
\includegraphics[trim=50 20 50 20,clip,height=\CurveHeight]{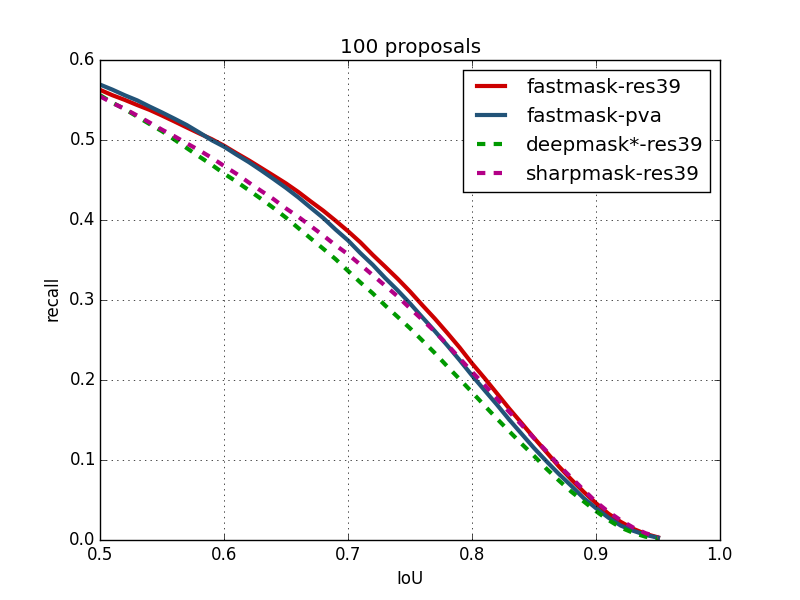} &
\includegraphics[trim=50 20 50 20,clip,height=\CurveHeight]{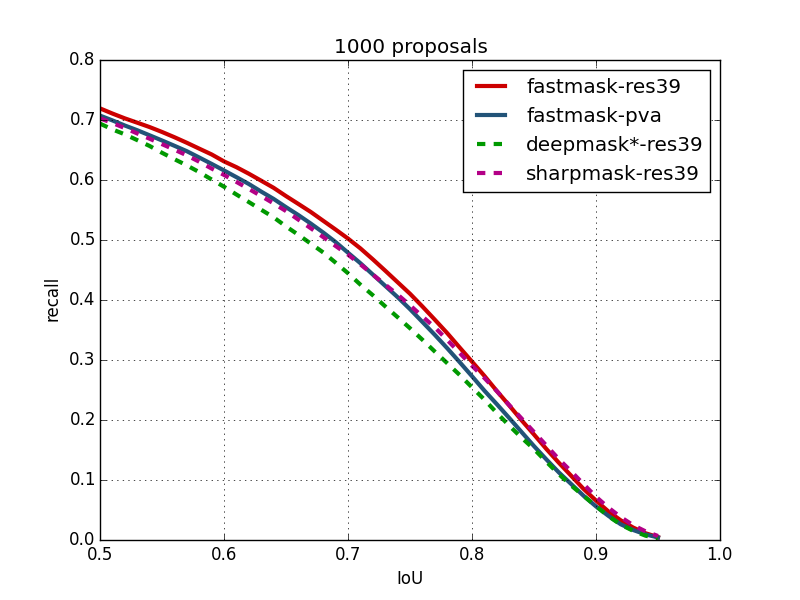} \\

\end{tabular}
\caption{ Proposal recall curve. (a-c) show detailed box proposal recall while (d-e) show detailed segment proposal recall. }
\vspace*{-0.1in}
\label{fig:result:plot}
\end{figure*}

We analyze and evaluate our network on MS COCO benchmark, which contains 80k training images and a total of nearly 500k instance annotations. Following the experimental setting of ~\cite{pinheiro_nips15,pinheiro_eccv16,dai_eccv16}, we report our result on the first 5k COCO validation images. We use another non-overlapped 5k images for validation.

\noindent{\textbf{Metrics}}. We measure the mask accuracy by Intersection over Union(IoU) between predicted mask and ground truth annotation. As average recall correlates well with object proposal quality~\cite{hosang_pami16}, we summarize Average Recall (AR) between IoU 0.5 and 0.95 for a fixed number N of proposals, denoted as "AR@N" in order to measure the performance of algorithms. (We use N equals to 10, 100 and 1000)

\noindent{\textbf{Scales}}. As COCO dataset contains objects in a wide range of scales, a more fine-grained evaluation tends to measures metrics with regards to object scales. Practically, objects are divided into three groups according to their pixel areas $a$: small ($a<32^2$), medium ($32^2<a<96^2$), large ($a>96^2$). In our experiments, we denote the metrics for different scales by adding superscripts $S$, $M$, $L$ respectfully.

\noindent{\textbf{Methods}}. By default, we compare our method with recent state-of-the-arts for segment proposal, including \textit{{DeepMask}}~\cite{pinheiro_nips15}, \textit{{SharpMask}}~\cite{pinheiro_eccv16} and \textit{{InstanceFCN}}~\cite{dai_eccv16}. Note that we also provide results from a revised DeepMask architecture from ~\cite{pinheiro_eccv16}, denoted as \textit{{DeepMask$^*$}}. Different from original DeepMask, it is implemented based on 39-layer residual net with a revised head component. These methods not only achieve good Average Recall but also provide strong efficiency during inference.

Our network is general and could be plug-in to different body networks. In our experiments, we adopt 39-layer Residual Net~\cite{he_cvpr16} for best accuracy as well as fair comparison and PvaNet~\cite{kim_arxiv16} for best efficiency.

\subsection{ Comparison with State-of-the-art Methods }
\label{sec:exp:state-of-the-art}

Table~\ref{tab:exp:coco} compares the performance of our FastMask to other state-of-the-art methods. We report results on both bounding box and segment proposals (by deriving a tight bounding box from a mask proposal). Here we do not include the SharpMaskZoom$^2$ result because they use images with extra scales (2 \textasciicircum 1/2 larger) to obtain superior performance.

We compare our two-stream FastMask with all those image pyramid based methods since our one-stream network does not contain the same density in its feature pyramid. To address the influence of feature scale density to performance as well as efficiency, we conduct separate controlled experiments in Section~\ref{sec:exp:speed}.

\noindent{\textbf{Quantitative evaluation}}.
According to Table~\ref{tab:exp:coco}, we outperform all state-of-the-art methods in bounding-box proposal by a large margin and obtain very competitive results with segmentation proposals (outperform all methods on AR@10 and AR@1k, and show competitive performance on AR@100). It is worth noting that our two-stream network significantly improves the box proposal quality comparing to all other methods, which provides a guidance on its potential for bbox-based object detection. Our two-stream FastMask model with 39-layers Resnet achieves approximately 18\%, 11\%, 8\% relative improvement on  AR@10, AR@100, AR@1k metrics respectively, over previous best SharpMask model. In order to give a better picture of our proposal quality, we plot the recall versus IoU threshold for different of segmentation proposals in COCO dataset as Figure~\ref{fig:result:plot}. There is a clear gap in the plot, which indicates that FastMask produce better mask quality overall.

While obtaining superior performance, our method also yields better efficiency than all image pyramid based approaches. We did some controlled experiments and report the speed/performance in Section~\ref{sec:exp:speed}.

\noindent{\textbf{Qualitative visualization}}.
We visualize some results in Figure~\ref{fig:result:viz} showing exemplars on which our method improves over baselines. Generally, we observe that our method is more robust to scale variance and invariant to noisy background. Not like SharpMask, FastMask does not perform any mask refinement at all. It is possible to further boost mask quality by leveraging mask refinement.


\begin{figure*}[bp]
\small
\centering
\tabcolsep 1pt
\begin{tabular}{ccccc|ccccc}
Image & DeepMask$*$ & SharpMask & FastMask & Ground Truth & Image & DeepMask$*$ & SharpMask & FastMask & Ground Truth \\
\includegraphics[width=\VizImageWidth,height=\VizImageHeightLong]{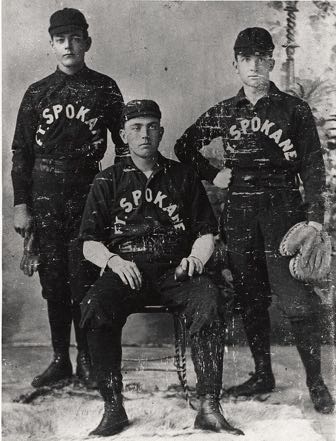} &
\includegraphics[width=\VizImageWidth,height=\VizImageHeightLong]{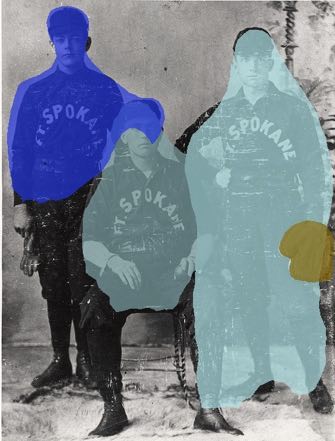} &
\includegraphics[width=\VizImageWidth,height=\VizImageHeightLong]{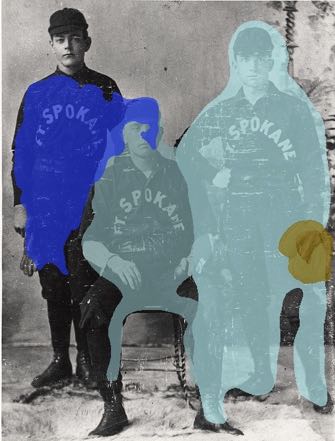} &
\includegraphics[width=\VizImageWidth,height=\VizImageHeightLong]{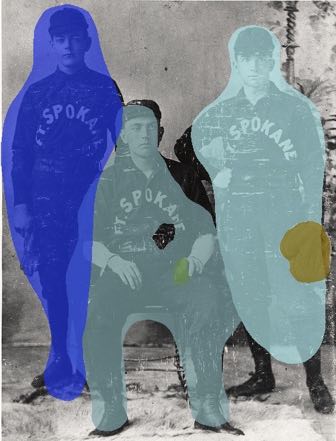} &
\includegraphics[width=\VizImageWidth,height=\VizImageHeightLong]{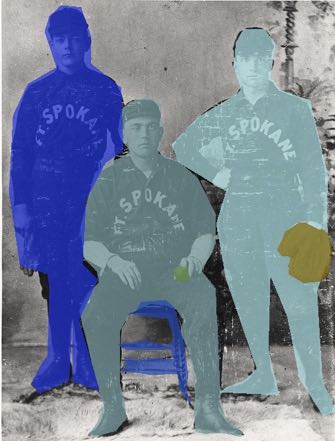} &

\includegraphics[width=\VizImageWidth,height=\VizImageHeightLong]{files/coco/02} &
\includegraphics[width=\VizImageWidth,height=\VizImageHeightLong]{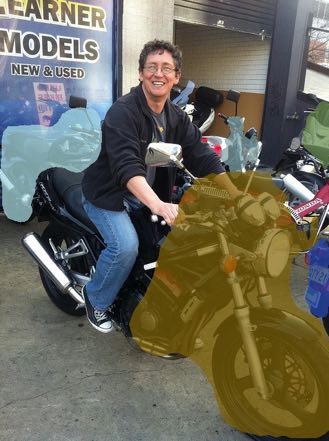} &
\includegraphics[width=\VizImageWidth,height=\VizImageHeightLong]{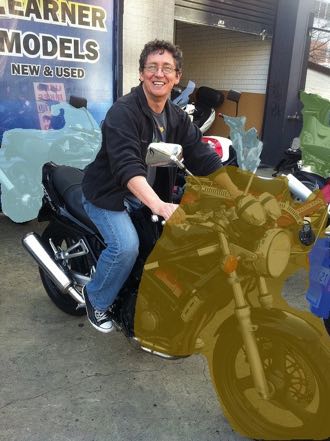} &
\includegraphics[width=\VizImageWidth,height=\VizImageHeightLong]{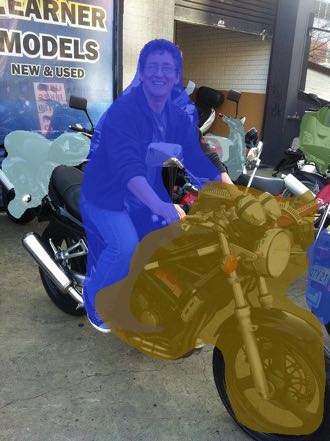} &
\includegraphics[width=\VizImageWidth,height=\VizImageHeightLong]{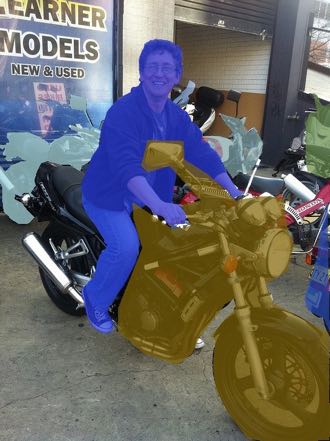} \\

\includegraphics[width=\VizImageWidth,height=\VizImageHeightMiddle]{files/coco/03} &
\includegraphics[width=\VizImageWidth,height=\VizImageHeightMiddle]{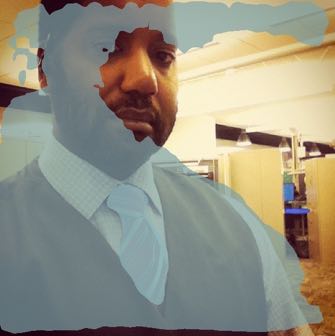} &
\includegraphics[width=\VizImageWidth,height=\VizImageHeightMiddle]{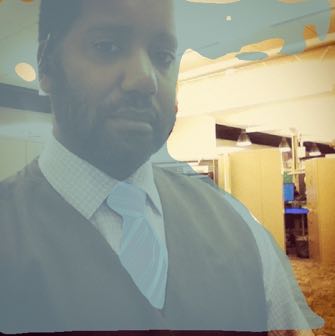} &
\includegraphics[width=\VizImageWidth,height=\VizImageHeightMiddle]{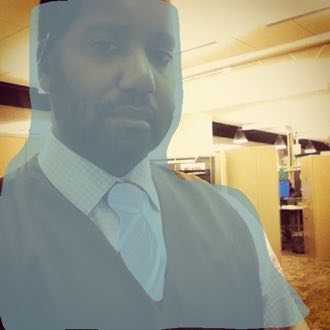} &
\includegraphics[width=\VizImageWidth,height=\VizImageHeightMiddle]{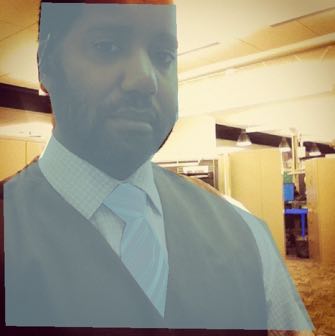} &

\includegraphics[width=\VizImageWidth,height=\VizImageHeightMiddle]{files/coco/04} &
\includegraphics[width=\VizImageWidth,height=\VizImageHeightMiddle]{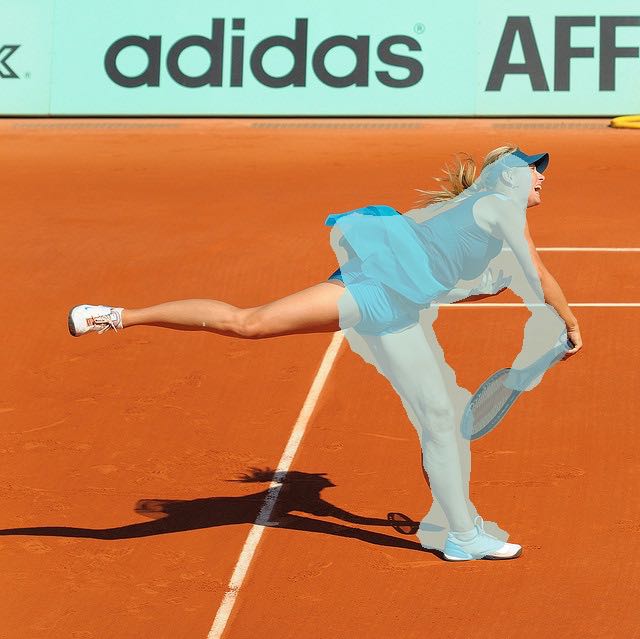} &
\includegraphics[width=\VizImageWidth,height=\VizImageHeightMiddle]{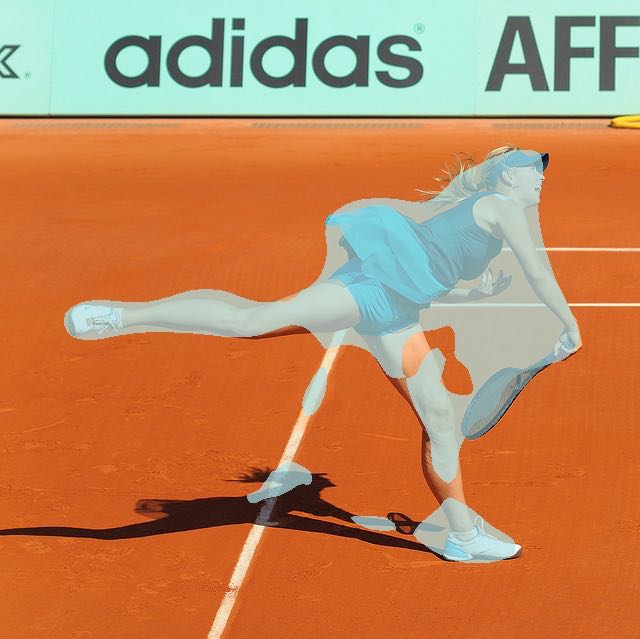} &
\includegraphics[width=\VizImageWidth,height=\VizImageHeightMiddle]{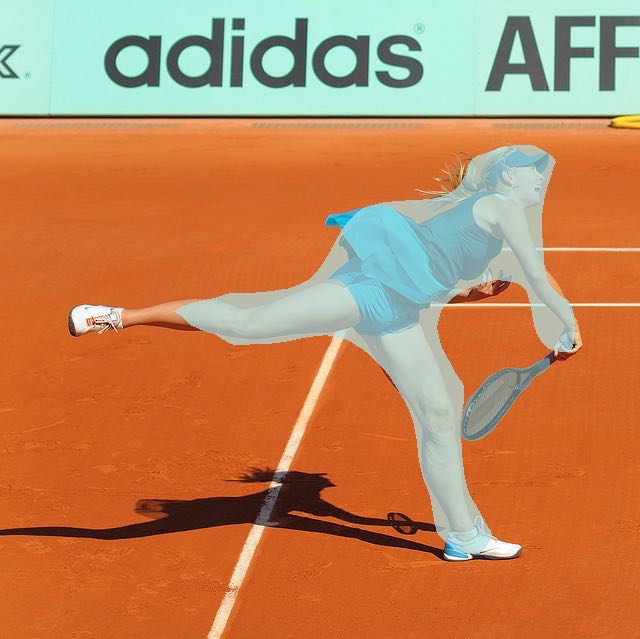} &
\includegraphics[width=\VizImageWidth,height=\VizImageHeightMiddle]{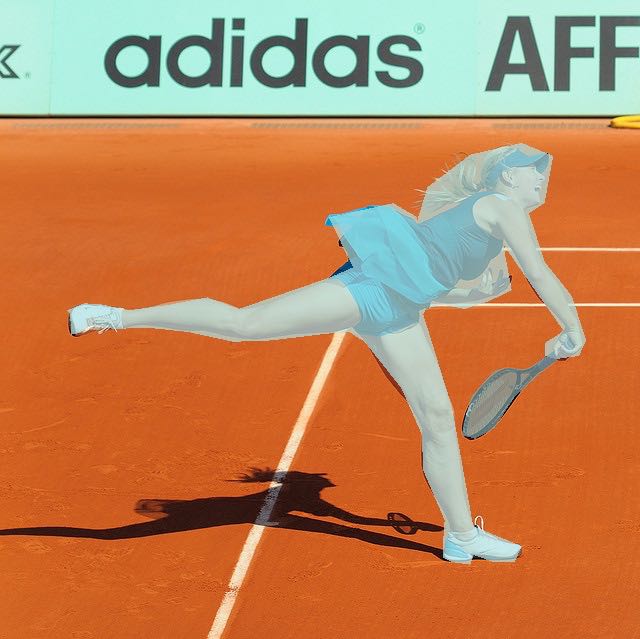} \\

\includegraphics[width=\VizImageWidth,height=\VizImageHeightLong]{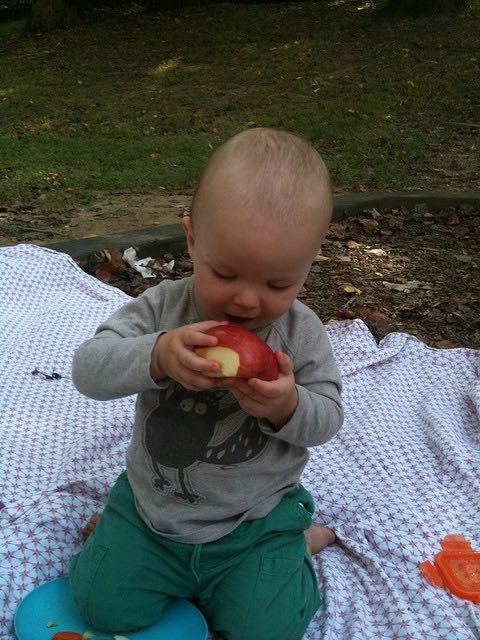} &
\includegraphics[width=\VizImageWidth,height=\VizImageHeightLong]{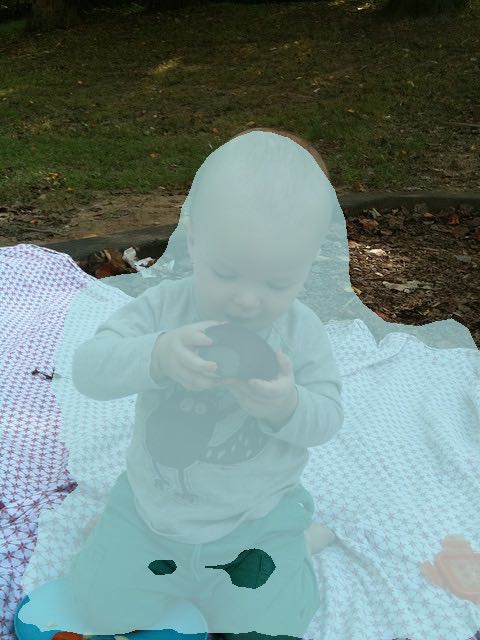} &
\includegraphics[width=\VizImageWidth,height=\VizImageHeightLong]{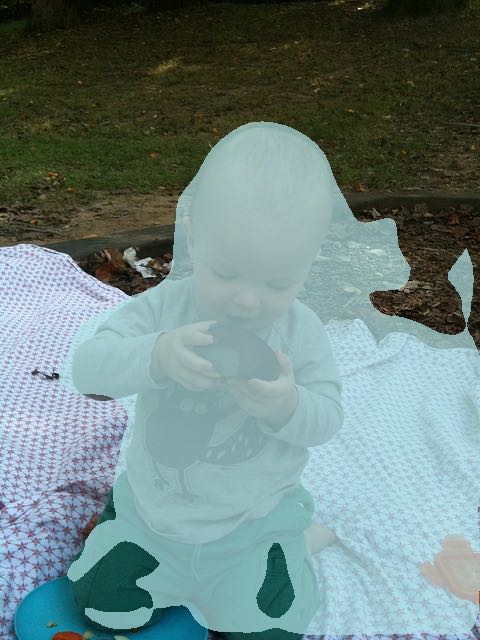} &
\includegraphics[width=\VizImageWidth,height=\VizImageHeightLong]{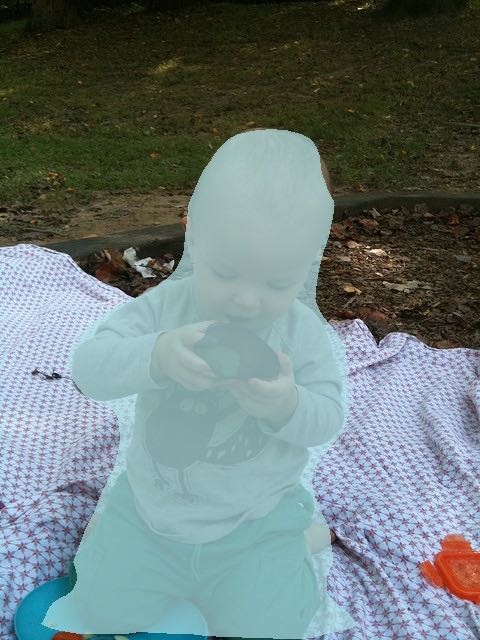} &
\includegraphics[width=\VizImageWidth,height=\VizImageHeightLong]{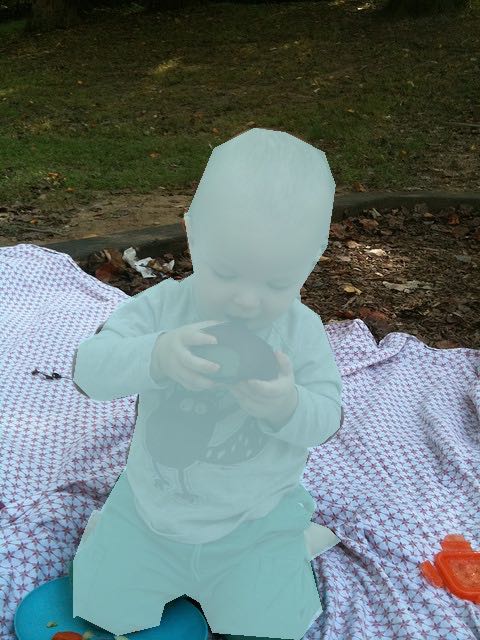} &

\includegraphics[width=\VizImageWidth,height=\VizImageHeightLong]{files/coco/06} &
\includegraphics[width=\VizImageWidth,height=\VizImageHeightLong]{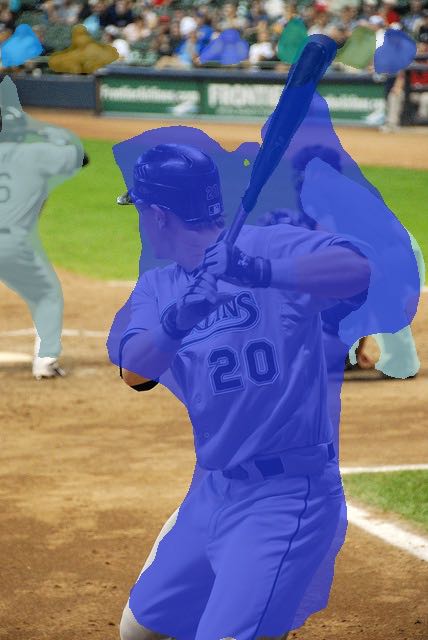} &
\includegraphics[width=\VizImageWidth,height=\VizImageHeightLong]{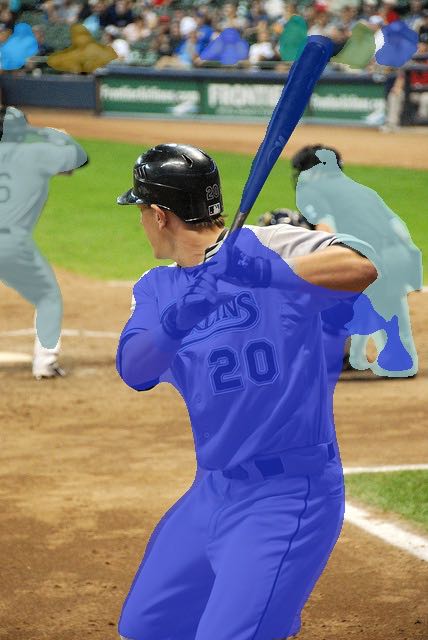} &
\includegraphics[width=\VizImageWidth,height=\VizImageHeightLong]{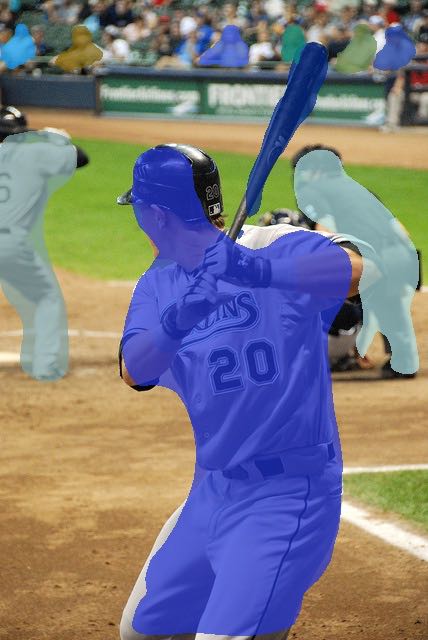} &
\includegraphics[width=\VizImageWidth,height=\VizImageHeightLong]{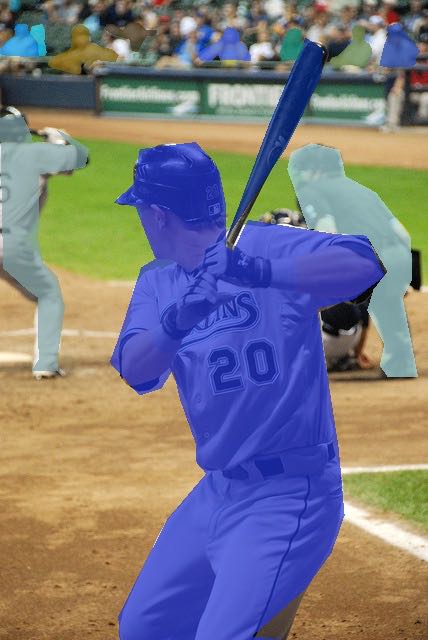} \\

\includegraphics[width=\VizImageWidth,height=\VizImageHeightLong]{files/coco/07} &
\includegraphics[width=\VizImageWidth,height=\VizImageHeightLong]{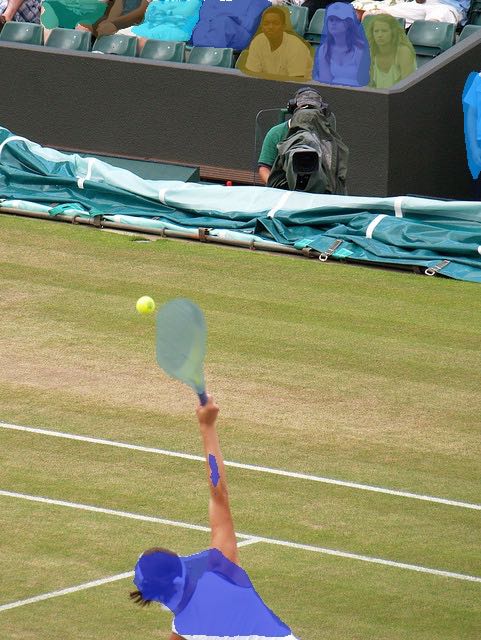} &
\includegraphics[width=\VizImageWidth,height=\VizImageHeightLong]{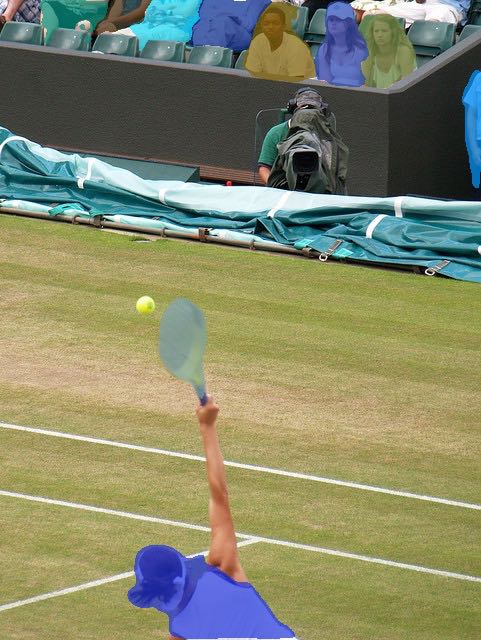} &
\includegraphics[width=\VizImageWidth,height=\VizImageHeightLong]{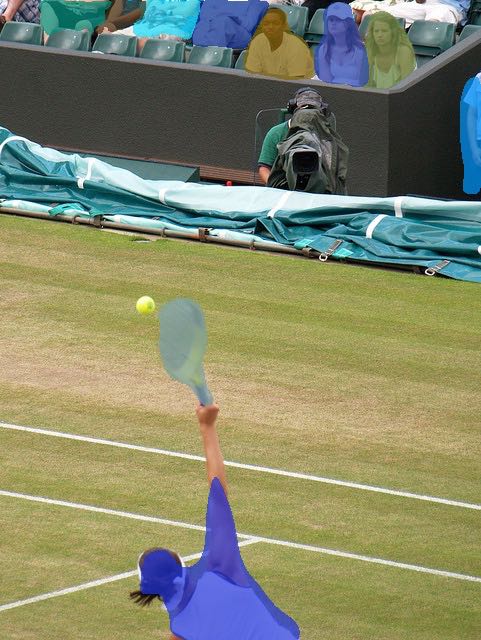} &
\includegraphics[width=\VizImageWidth,height=\VizImageHeightLong]{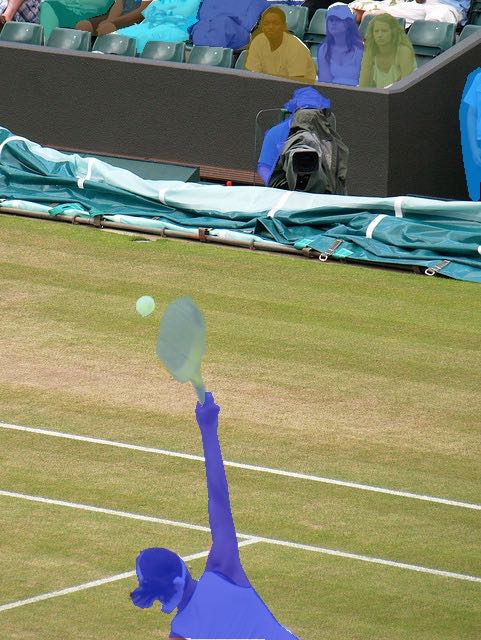} &

\includegraphics[width=\VizImageWidth,height=\VizImageHeightLong]{files/coco/08} &
\includegraphics[width=\VizImageWidth,height=\VizImageHeightLong]{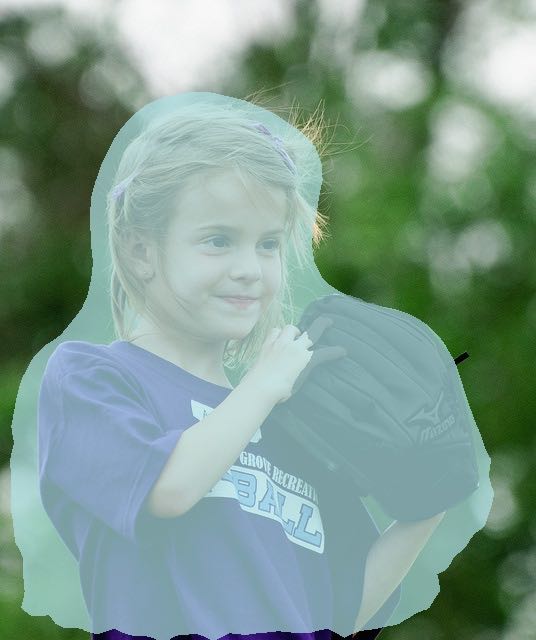} &
\includegraphics[width=\VizImageWidth,height=\VizImageHeightLong]{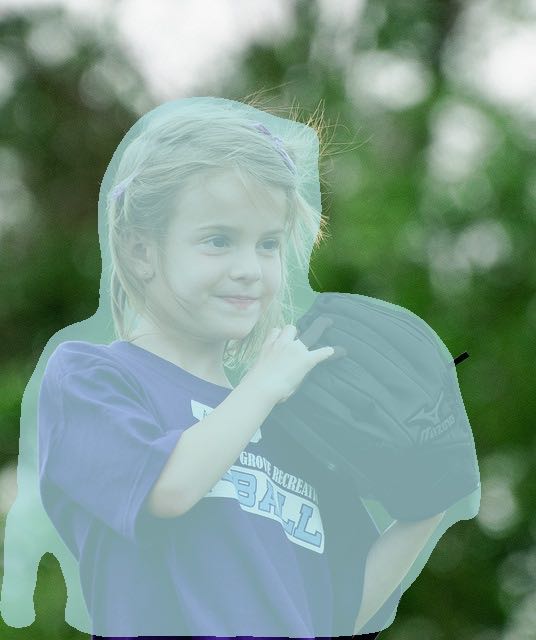} &
\includegraphics[width=\VizImageWidth,height=\VizImageHeightLong]{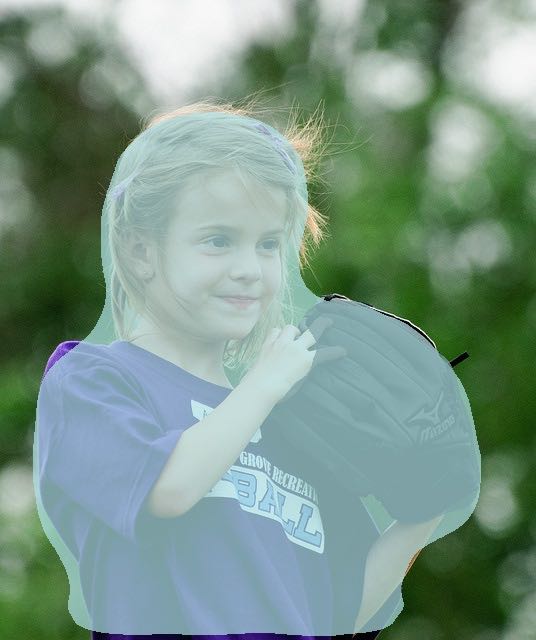} &
\includegraphics[width=\VizImageWidth,height=\VizImageHeightLong]{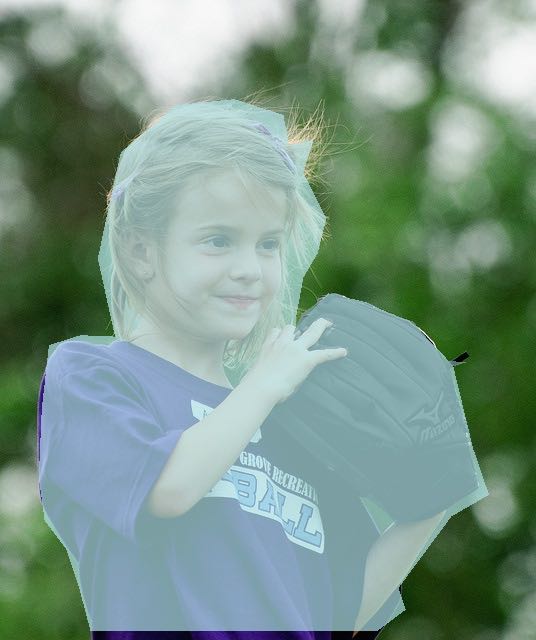} \\

\includegraphics[width=\VizImageWidth,height=\VizImageHeightMiddle]{files/coco/09} &
\includegraphics[width=\VizImageWidth,height=\VizImageHeightMiddle]{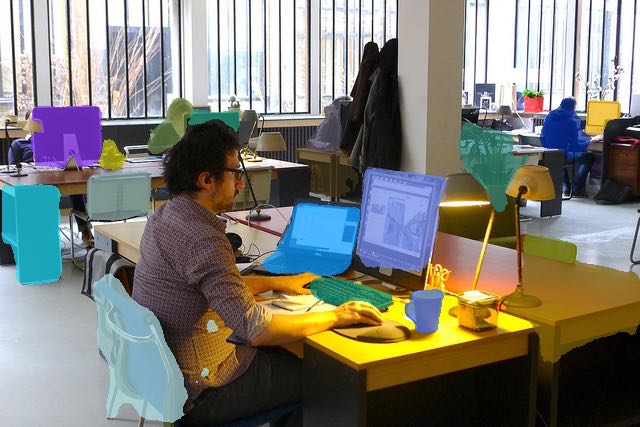} &
\includegraphics[width=\VizImageWidth,height=\VizImageHeightMiddle]{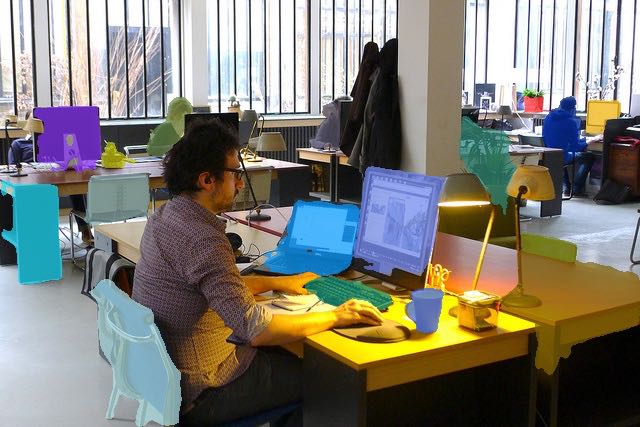} &
\includegraphics[width=\VizImageWidth,height=\VizImageHeightMiddle]{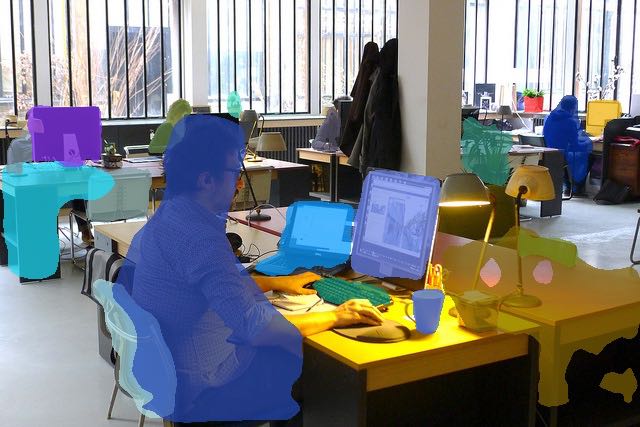} &
\includegraphics[width=\VizImageWidth,height=\VizImageHeightMiddle]{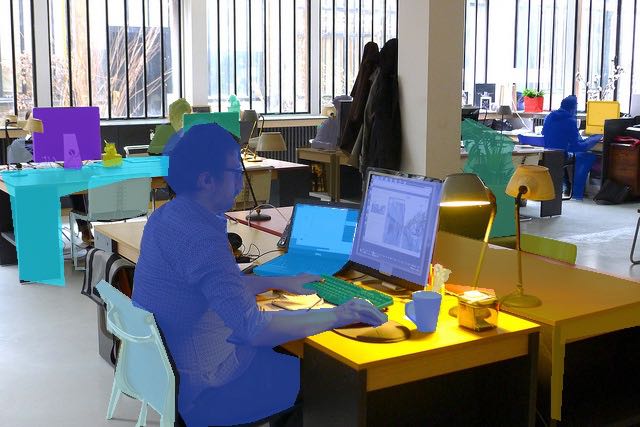} &

\includegraphics[width=\VizImageWidth,height=\VizImageHeightMiddle]{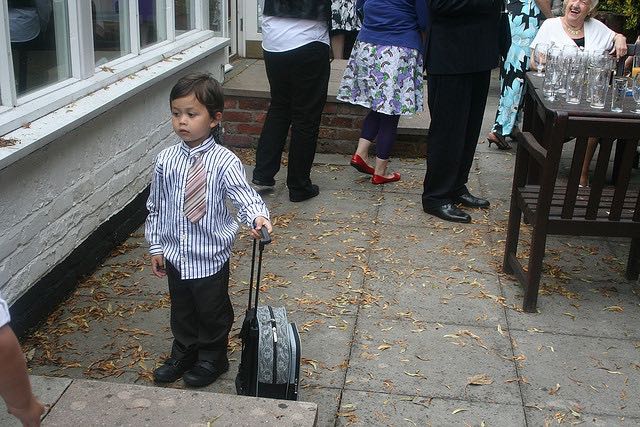} &
\includegraphics[width=\VizImageWidth,height=\VizImageHeightMiddle]{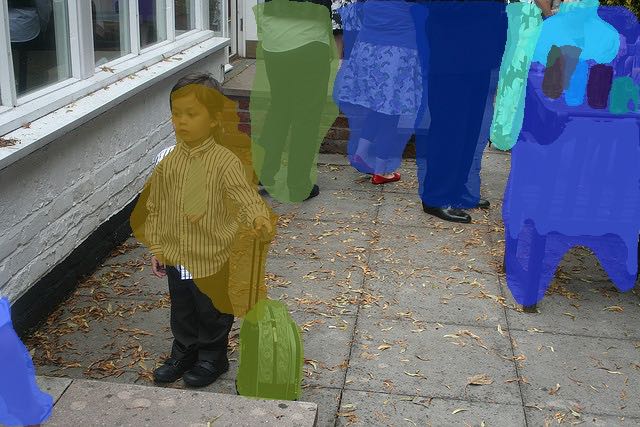} &
\includegraphics[width=\VizImageWidth,height=\VizImageHeightMiddle]{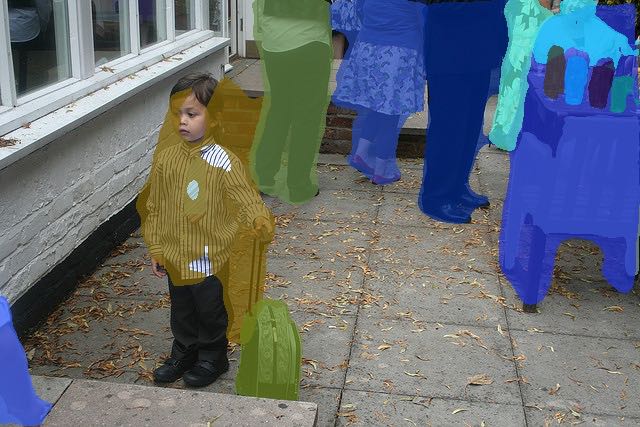} &
\includegraphics[width=\VizImageWidth,height=\VizImageHeightMiddle]{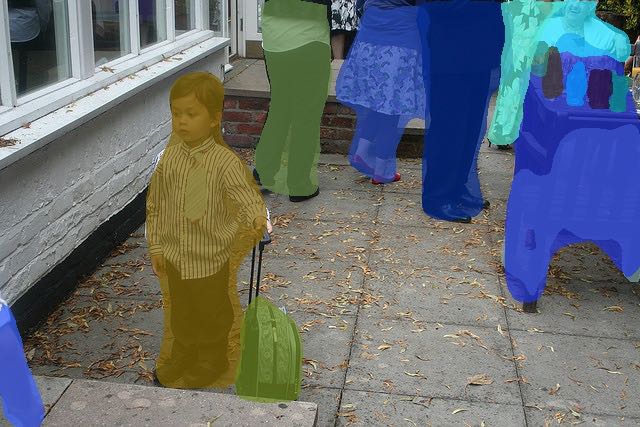} &
\includegraphics[width=\VizImageWidth,height=\VizImageHeightMiddle]{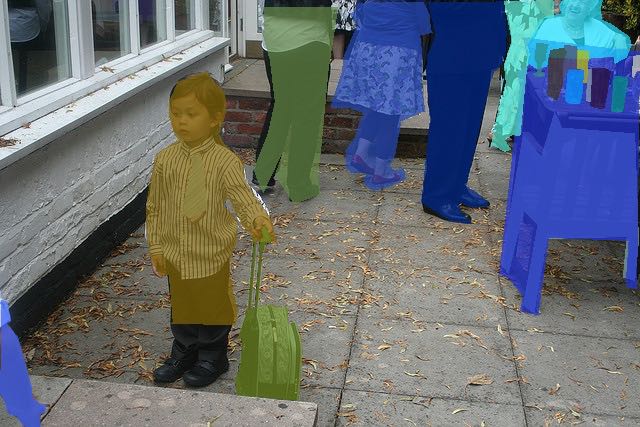} \\

\includegraphics[width=\VizImageWidth,height=\VizImageHeightMiddle]{files/coco/11} &
\includegraphics[width=\VizImageWidth,height=\VizImageHeightMiddle]{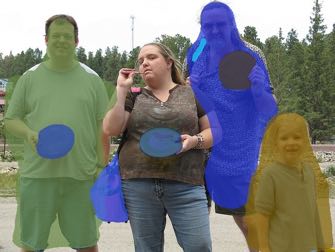} &
\includegraphics[width=\VizImageWidth,height=\VizImageHeightMiddle]{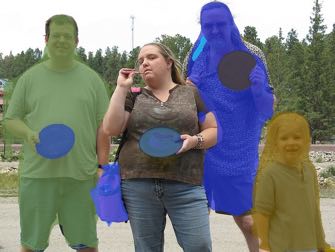} &
\includegraphics[width=\VizImageWidth,height=\VizImageHeightMiddle]{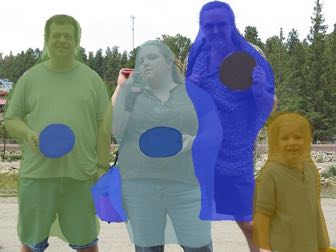} &
\includegraphics[width=\VizImageWidth,height=\VizImageHeightMiddle]{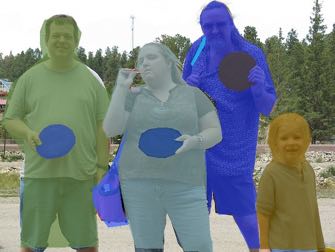} &

\includegraphics[width=\VizImageWidth,height=\VizImageHeightMiddle]{files/coco/12} &
\includegraphics[width=\VizImageWidth,height=\VizImageHeightMiddle]{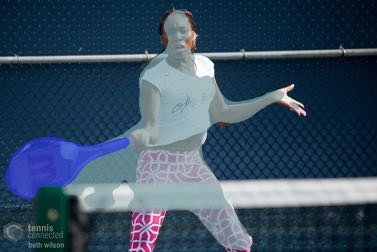} &
\includegraphics[width=\VizImageWidth,height=\VizImageHeightMiddle]{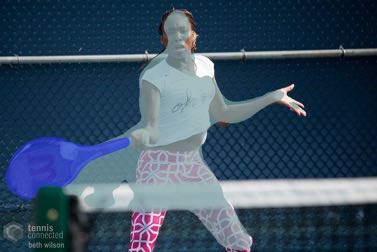} &
\includegraphics[width=\VizImageWidth,height=\VizImageHeightMiddle]{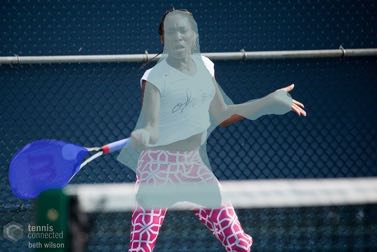} &
\includegraphics[width=\VizImageWidth,height=\VizImageHeightMiddle]{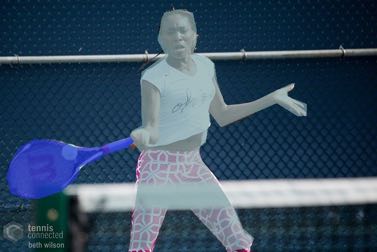} \\

\includegraphics[width=\VizImageWidth,height=\VizImageHeightMiddle]{files/coco/13} &
\includegraphics[width=\VizImageWidth,height=\VizImageHeightMiddle]{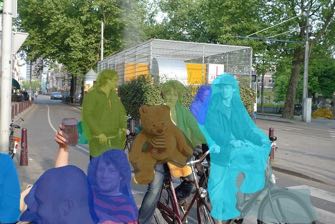} &
\includegraphics[width=\VizImageWidth,height=\VizImageHeightMiddle]{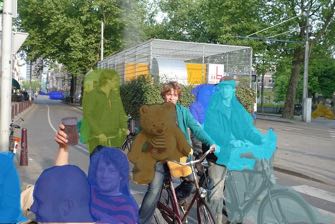} &
\includegraphics[width=\VizImageWidth,height=\VizImageHeightMiddle]{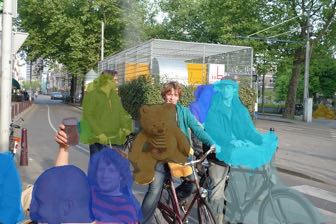} &
\includegraphics[width=\VizImageWidth,height=\VizImageHeightMiddle]{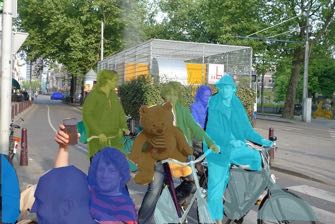} &

\includegraphics[width=\VizImageWidth,height=\VizImageHeightMiddle]{files/coco/14} &
\includegraphics[width=\VizImageWidth,height=\VizImageHeightMiddle]{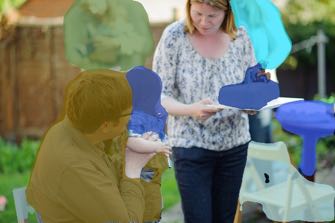} &
\includegraphics[width=\VizImageWidth,height=\VizImageHeightMiddle]{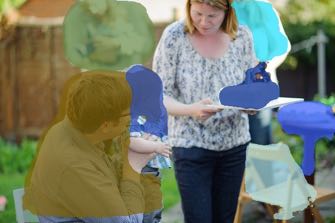} &
\includegraphics[width=\VizImageWidth,height=\VizImageHeightMiddle]{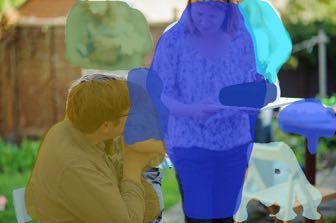} &
\includegraphics[width=\VizImageWidth,height=\VizImageHeightMiddle]{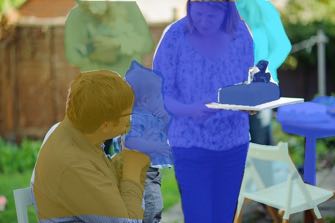} \\

\end{tabular}
\caption{Visualization of the object candidate segmentation results on sample MS COCO images. We compare our FastMask with DeepMask$*$~\cite{pinheiro_eccv16} and SharpMask~\cite{pinheiro_eccv16}. We also show  the origin images and the ground-truth annotations for reference.}
\label{fig:sup:viz:coco}
\end{figure*}

\subsection{ Efficiency Study }
\label{sec:exp:speed}

In this section, we evaluate two threads to support our argument that FastMask outperforms image pyramid methods on both efficiency and performance. On the first thread, we provide experimental results on DeepMask and SharpMask, with restriction on the scale density of their image pyramids. We construct a fair environment that both these methods and our method take equivalently many scales and evaluate both inference speed and performance. On the other thread, we provide the performance and speed of state-of-the-art methods and compare our best model as well as fastest model to them.

\begin{table}[tp]
\small
\tabcolsep 5pt
\centering
\begin{tabular}{ cc|cc|c }
Method & Scales & AR@10 & AR@100  & Speed\\ \hlineB{3}
{DeepMask$^*$~\cite{pinheiro_eccv16}} &8 & 14.3 & 27.3 & 0.45s   \\
{DeepMask$^*$~\cite{pinheiro_eccv16}} &4 & 11.3 & 22.2 & 0.24s   \\
\hline
{FastMask} &8 & 16.9 & 31.3 & 0.26s\\
{FastMask} &4 & 13.3 & 26.6 & 0.14s\\
\hline
\end{tabular}
\vspace{0.05in}
\caption{Trade-off between scale density and performance.}
\vspace{-0.15in}
\label{tab:result:scale}
\end{table}

\noindent{\textbf{Trade-off scale density with speed}}.
We conduct a fair study to analyze the trade-off by decreasing scale density. In the DeepMaskZoom$^*$ and SharpMaskZoom, they inference on images scaled from 2\textasciicircum [-2.5, -2.0, -1.5, -1.0, -0.5, 0, 0.5, 1] to obtain superior performance on a diverse range of object segments. This is similar to our two-stream network, where we input a image up-sampled by two. To improve the inference efficiency, we made a trade-off in scale density by reducing our network to one-stream without re-training, which is identical to reduce scale density for DeepMaskZoom$^*$ and SharpMaskZoom to 2\textasciicircum [-2.5, -1.5, -0.5, 0.5].

Figure~\ref{tab:result:scale} illustrates the performance degradation and efficiency increase with scale density trade-off. We measure only AR@10 and AR@100 as a sparse scale density leads to less total proposal number. These controlled experiments are tested using NVIDIA Titan X GPU. We do multiple runs and average their time to obtain an estimation of runtime speed. Our method achieves to preserve the best performance while increase the inference speed by almost 2$\times$. Note that retraining a network with reduced scale density can boost up performance.

\noindent{\textbf{Speed evaluation}}.
We evaluate the inference speed of all state-of-the-art methods. Two variant of our models, our most effective model (FastMask-acc) and most efficient model (FastMask-fast), are reported. Our most effective model takes a two-stream structure with 39-layer ResNet; Our fastest model takes a one-stream structure with PvaNet~\cite{kim_arxiv16}, which is light-weight and fast.

\begin{table}[t]
\small
\tabcolsep 1pt
\centering
\begin{tabular}{ cc|ccc|c }
Method & Body Net \ & \ AR@10 & AR@100 & AR@1k \ & \ Speed\ \\ \hlineB{3}
{DeepMask~\cite{pinheiro_nips15}}       & VGG   & 12.6 & 24.5 & 33.1 & 1.60s \\
{DeepMask$^*$~\cite{pinheiro_eccv16}}   & Res39 & 14.1 & 25.8 & 33.6 & 0.46s \\
{SharpMask~\cite{pinheiro_eccv16}}      & Res39 & 15.4 & 27.8 & 36.0 & 0.76s \\
{SharpMaskZoom~\cite{pinheiro_eccv16}}  & Res39 & 15.6 & 30.4 & \second{40.1} & $\sim$1.5s   \\
{InstanceFCN~\cite{dai_eccv16}}         & VGG   & 16.6 & \bst{31.7} & 39.2 & 1.50s  \\
\hline
{FastMask-acc}   & Res39 & \second{16.9} & \second{31.3} & \bst{40.6} & 0.26s\\
{FastMask-fast}  & Pva   & \bst{17.2}    & 29.4          & 36.4 & 0.07s\\
\hline
\end{tabular}
\vspace{0.05in}
\caption{Speed Study with state-of-the-art methods.}
\vspace{-0.2in}
\label{tab:result:speed}
\end{table}

Figure~\ref{tab:result:speed} compare our best and fastest model with other networks. Our best model produces superior proposal performance while preserving good efficiency. With slight trade-off in performance, our fastest model obtains almost real-time efficiency ($\sim$13 FPS by NVIDIA Titan X Maxwell).

\section{ Conclusion }

In this paper we present an innovative framework, \ie FastMask, for efficient segment-based object proposal. Instead of building pyramid of input image, FastMask learns to encode feature pyramid by a neck module, and performs one-shot training and inference. Along with with process, a scale-tolerant head module is proposed to highlight the foreground object from its background noises, havesting a significant better segmentation accuracy. On MS COCO benchmark, FastMask outperforms all state-of-the-art segment proposal methods in average recall while keeping several times faster. More impressively, with a slight trade-off in accuracy, FastMast can segment objects in nearly real time ($\sim$13 fps) with images at 800$\times$600 resolution. As an effective and efficient segment proposal method, FastMask is believed to have great potentials in other tasks.

\section{ Acknowledgement }

H.X. Hu and F. Sha are partially supported by  NSF IIS-1065243, 1451412, 1513966, 1208500, CCF-1139148, a Google Research Award, an Alfred. P. Sloan Research Fellowship and ARO\#  W911NF-15-1-0484

{\small
\bibliographystyle{ieee}
\bibliography{ref}
}

\end{document}